\documentclass[preprint,12pt]{elsarticle}

\usepackage{amsmath,amssymb,amsfonts}
\usepackage{bm}
\usepackage{mathtools}
\usepackage{graphicx}
\usepackage{xcolor}
\usepackage{booktabs}
\usepackage{siunitx}
\usepackage{enumitem}
\usepackage{algorithm}
\usepackage{algpseudocode}
\usepackage{caption}
\usepackage{subcaption}
\usepackage{xspace}

\usepackage{hyperref}
\usepackage{url}

\makeatletter
\DeclareRobustCommand{\path}[1]{\url{#1}} 
\makeatother

\hypersetup{
    colorlinks = true,
    linkcolor  = blue,
    citecolor  = blue,
    urlcolor   = blue
}

\def\XS{\xspace}
\DeclareMathAlphabet{\mathb}{OML}{cmm}{b}{it}
\def\sbmm#1{\ensuremath{\boldsymbol{#1}}}          
\def\bm#1{\mbox{\boldmath $#1$}}

\def\Ab{{\sbm{A}}\XS}  \def\ab{{\sbm{a}}\XS}
\def\Bb{{\sbm{B}}\XS}  \def\bb{{\sbm{b}}\XS}
\def\Cb{{\sbm{C}}\XS}  \def\cb{{\sbm{c}}\XS}
\def\Db{{\sbm{D}}\XS}  \def\db{{\sbm{d}}\XS}
\def\Eb{{\sbm{E}}\XS}  \def\eb{{\sbm{e}}\XS}
\def\Fb{{\sbm{F}}\XS}  \def\fb{{\sbm{f}}\XS}
\def\Gb{{\sbm{G}}\XS}  \def\gb{{\sbm{g}}\XS}
\def\Hb{{\sbm{H}}\XS}  \def\hb{{\sbm{h}}\XS}
\def\Ib{{\sbm{I}}\XS}  \def\ib{{\sbm{i}}\XS}
\def\Jb{{\sbm{J}}\XS}  \def\jb{{\sbm{j}}\XS}
\def\Kb{{\sbm{K}}\XS}  \def\kb{{\sbm{k}}\XS}
\def\Lb{{\sbm{L}}\XS}  \def\lb{{\sbm{l}}\XS}
\def\Mb{{\sbm{M}}\XS}  \def\mb{{\sbm{m}}\XS}
\def\Nb{{\sbm{N}}\XS}  \def\nb{{\sbm{n}}\XS}
\def\Ob{{\sbm{O}}\XS}  \def\ob{{\sbm{o}}\XS}
\def\Pb{{\sbm{P}}\XS}  \def\pb{{\sbm{p}}\XS}
\def\Qb{{\sbm{Q}}\XS}  \def\qb{{\sbm{q}}\XS}
\def\Rb{{\sbm{R}}\XS}  \def\rb{{\sbm{r}}\XS}
\def\Sb{{\sbm{S}}\XS}  \def\sb{{\sbm{s}}\XS}
\def\Tb{{\sbm{T}}\XS}  \def\tb{{\sbm{t}}\XS}
\def\Ub{{\sbm{U}}\XS}  \def\ub{{\sbm{u}}\XS}
\def\Vb{{\sbm{V}}\XS}  \def\vb{{\sbm{v}}\XS}
\def\Wb{{\sbm{W}}\XS}  \def\wb{{\sbm{w}}\XS}
\def\Xb{{\sbm{X}}\XS}  \def\xb{{\sbm{x}}\XS}
\def\Yb{{\sbm{Y}}\XS}  \def\yb{{\sbm{y}}\XS}
\def\Zb{{\sbm{Z}}\XS}  \def\zb{{\sbm{z}}\XS}

\def\Ac{{\scu{A}}\XS}   
\def\Bc{{\scu{B}}\XS}   
\def\Cc{{\scu{C}}\XS}   
\def\Dc{{\scu{D}}\XS}   
\def\Ec{{\scu{E}}\XS}   
\def\Fc{{\scu{F}}\XS}   
\def\Gc{{\scu{G}}\XS}   
\def\Hc{{\scu{H}}\XS}   
\def\Ic{{\scu{I}}\XS}   
\def\Jc{{\scu{J}}\XS}   
\def\Kc{{\scu{K}}\XS}   
\def\Lc{{\scu{L}}\XS}   
\def\Mc{{\scu{M}}\XS}   
\def\Nc{{\scu{N}}\XS}   
\def\Oc{{\scu{O}}\XS}   
\def\Pc{{\scu{P}}\XS}   
\def\Qc{{\scu{Q}}\XS}   
\def\Rc{{\scu{R}}\XS}   
\def\Sc{{\scu{S}}\XS}   
\def\Tc{{\scu{T}}\XS}   
\def\Uc{{\scu{U}}\XS}   
\def\Vc{{\scu{V}}\XS}   
\def\Wc{{\scu{W}}\XS}   
\def\Xc{{\scu{X}}\XS}   
\def\Yc{{\scu{Y}}\XS}   
\def\Zc{{\scu{Z}}\XS}   

  \def\rv{{\sbv{r}}\XS}

\def\alphab{{\sbmm{\alpha}}\XS}
\def\betab{{\sbmm{\beta}}\XS}
\def\gammab{{\sbmm{\gamma}}\XS}      
\def\deltab{{\sbmm{\delta}}\XS}      \def\Deltab{{\sbmm{\Delta}}\XS}
\def\epsilonb{{\sbmm{\epsilon}}\XS}

\def\etab{{\sbmm{\eta}}\XS}
\def\thetab{{\sbmm{\theta}}\XS}      \def\Thetab{{\sbmm{\Theta}}\XS}

\def\lambdab{{\sbmm{\lambda}}\XS}     \def\Lambdab{{\sbmm{\Lambda}}\XS}
\def\mub{{\sbmm{\mu}}\XS}
\def\nub{{\sbmm{\nu}}\XS}
\def\xib{{\sbmm{\xi}}\XS}

\def\rhob{{\sbmm{\rho}}\XS}

\def\sigmab{{\sbmm{\sigma}}\XS}      	\def\Sigmab{{\sbmm{\Sigma}}\XS}
   
\def\phib{{\sbmm{\phi}}\XS}        		\def\Phib{{\sbmm{\Phi}}\XS}

\def\chib{{\sbmm{\chi}}\XS}
\def\psib{{\sbmm{\psi}}\XS}        		\def\Psib{{\sbmm{\Psi}}\XS}
\def\omegab{{\sbmm{\omega}}\XS}      	
\def\taub{{\sbmm{\tau}}\XS}
   	
\def\zetab{{\sbmm{\zeta}}\XS}

  \def\fh{{\widehat{f}}\XS}

\def\Mh{{\widehat{M}}\XS}

  \def\qh{{\widehat{q}}\XS}
  \def\rh{{\widehat{r}}\XS}
  
  \def\th{{\widehat{t}}\XS}
  
  \def\vh{{\widehat{v}}\XS}
  \def\wh{{\widehat{w}}\XS}
  \def\xh{{\widehat{x}}\XS}
  
  \def\zh{{\widehat{z}}\XS}

  \def\qt{{\widetilde{q}}\XS}
  \def\zt{{\widetilde{z}}\XS}

\def\Abt{{\widetilde{\Ab}}\XS}  \def\abt{{\widetilde{\ab}}\XS}

\def\Dbt{{\widetilde{\Db}}\XS}  
  
  \def\fbt{{\widetilde{\fb}}\XS}

  \def\kbt{{\widetilde{\kb}}\XS}

\def\Zbt{{\widetilde{\Zb}}\XS}  \def\zbt{{\widetilde{\zb}}\XS}

\def\Abh{{\widehat{\Ab}}\XS}  \def\abh{{\widehat{\ab}}\XS}
\def\Bbh{{\widehat{\Bb}}\XS}  
  
  \def\dbh{{\widehat{\db}}\XS}
  
\def\Fbh{{\widehat{\Fb}}\XS}  \def\fbh{{\widehat{\fb}}\XS}
  \def\gbh{{\widehat{\gb}}\XS}
  \def\hbh{{\widehat{\hb}}\XS}

\def\Pbh{{\widehat{\Pb}}\XS}  
  \def\qbh{{\widehat{\qb}}\XS}
  
  \def\sbh{{\widehat{\sb}}\XS}
  \def\tbh{{\widehat{\tb}}\XS}
  \def\ubh{{\widehat{\ub}}\XS}
\def\Vbh{{\widehat{\Vb}}\XS}  \def\vbh{{\widehat{\vb}}\XS}
  
  \def\xbh{{\widehat{\xb}}\XS}
  \def\ybh{{\widehat{\yb}}\XS}
  \def\zbh{{\widehat{\zb}}\XS}

\def\alphah{{\widehat{\alpha}}\XS}
\def\betah{{\widehat{\beta}}\XS}

\def\thetah{{\widehat{\theta}}\XS}

\def\lambdah{{\widehat{\lambda}}\XS}     
\def\muh{{\widehat{\mu}}\XS}

\def\rhoh{{\widehat{\rho}}\XS}

\def\sigmah{{\widehat{\sigma}}\XS}

\def\tauh{{\widehat{\tau}}\XS}

\def\alphat{{\widetilde{\alpha}}\XS}
\def\betat{{\widetilde{\beta}}\XS}
\def\gammat{{\widetilde{\gamma}}\XS}

\def\etat{{\widetilde{\eta}}\XS}
\def\thetat{{\widetilde{\theta}}\XS}

\def\mut{{\widetilde{\mu}}\XS}

      	\def\Sigmat{{\widetilde{\Sigma}}\XS}

\def\taut{{\widetilde{\tau}}\XS}
   	
\def\zetat{{\widetilde{\zeta}}\XS}

\def\alphabh{{\widehat{\alphab}}\XS}
\def\betabh{{\widehat{\betab}}\XS}

\def\etabh{{\widehat{\etab}}\XS}
\def\thetabh{{\widehat{\thetab}}\XS}      \def\Thetabh{{\widehat{\Thetab}}\XS}

\def\lambdabh{{\widehat{\lambdab}}\XS}     
\def\mubh{{\widehat{\mub}}\XS}

\def\rhobh{{\widehat{\rhob}}\XS}

\def\sigmabh{{\widehat{\sigmab}}\XS}      	\def\Sigmabh{{\widehat{\Sigmab}}\XS}

\def\thetabt{{\widetilde{\thetab}}\XS}

\def\mubt{{\widetilde{\mub}}\XS}
\def\nubt{{\widetilde{\nub}}\XS}

      	\def\Sigmabt{{\widetilde{\Sigmab}}\XS}

\def\blue#1{{\color{blue}#1}}
\def\red#1{{\color{red}#1}} 
\def\green#1{{\color{green}#1}} 

\def\rF{{\red{F}}}  
\def\rFb{{\red{\Fb}}}  \def\rfj{{\red{f_j}}}
\def\rZb{{\red{\Zb}}}  \def\rzj{{\red{z_j}}}

\def\rtaub{{\red{\taub}}}

\def\rvzj{\red{{v_z}_j}}

\def\bG{{\blue{G}}}  \def\bg{{\blue{g}}}

\def\rAb{{\red{\Ab}}}  \def\rab{{\red{\ab}}}
\def\rBb{{\red{\Bb}}}  \def\rbb{{\red{\bb}}}

\def\rFb{{\red{\Fb}}}  \def\rfb{{\red{\fb}}}
  \def\rgb{{\red{\gb}}}
\def\rHb{{\red{\Hb}}}  \def\rhb{{\red{\hb}}}

  \def\rmb{{\red{\mb}}}

  \def\rqb{{\red{\qb}}}

  \def\rtb{{\red{\tb}}}
  \def\rub{{\red{\ub}}}
\def\rVb{{\red{\Vb}}}  \def\rvb{{\red{\vb}}}
  \def\rwb{{\red{\wb}}}

\def\rZb{{\red{\Zb}}}  \def\rzb{{\red{\zb}}}

\def\bFb{{\blue{\Fb}}}  \def\bfb{{\blue{\fb}}}
  \def\bgb{{\blue{\gb}}}
\def\bHb{{\blue{\Hb}}}

\def\rAbh{{\widehat{\rAb}}}  \def\rabh{{\widehat{\rab}}}
\def\rBbh{{\widehat{\rBb}}}

\def\rFbh{{\widehat{\rFb}}}  \def\rfbh{{\widehat{\rfb}}}
  \def\rgbh{{\widehat{\rgb}}}
  \def\rhbh{{\widehat{\rhb}}}

  \def\rmbh{{\widehat{\rmb}}}

\def\rPbh{{\widehat{\rPb}}}  
  \def\rqbh{{\widehat{\rqb}}}

  \def\rubh{{\widehat{\rub}}}
\def\rVbh{{\widehat{\rVb}}}  \def\rvbh{{\widehat{\rvb}}}
  \def\rwbh{{\widehat{\rwb}}}

  \def\rzbh{{\widehat{\rzb}}}

  \def\rfbt{{\widetilde{\rfb}}}

\def\rVbt{{\widetilde{\rVb}}}  \def\rvbt{{\widetilde{\rvb}}}

  \def\rzbt{{\widetilde{\rzb}}}

\def\ralphab{{\red{\alphab}}}
\def\rbetab{{\red{\betab}}}

\def\retab{{\red{\etab}}}
\def\rthetab{{\red{\thetab}}}

\def\rlambdab{{\red{\lambdab}}}     

\def\rnub{{\red{\nub}}}

      	\def\rSigmab{{\red{\Sigmab}}}
   
\def\rphib{{\red{\phib}}}

\def\rtaub{{\red{\taub}}}

\def\ralphabh{{\red{\alphabh}}}

\def\rthetabh{{\red{\thetabh}}}

      	\def\rSigmabh{{\red{\Sigmabh}}}

\def\rtaubh{{\red{\taubh}}}

\def\ralphabt{{\red{\alphabt}}}

\def\retabt{{\red{\etabt}}}
\def\rthetabt{{\red{\thetabt}}}

      	\def\rSigmabt{{\red{\Sigmabt}}}

\def\bdoc{\begin{document}}             \def\edoc{\end{document}}
\def\babs{\begin{abstract}}             \def\eabs{\end{abstract}}
\def\bcc{\begin{center}}                \def\ecc{\end{center}}
\def\ben{\begin{enumerate}}             \def\een{\end{enumerate}}
\def\bit{\begin{itemize}}               \def\eit{\end{itemize}}
\def\bpic{\begin{picture}}              \def\epic{\end{picture}}
\def\barr{\begin{array}}                \def\earr{\end{array}}
\def\btabu{\begin{tabular}}             \def\etabu{\end{tabular}}
\def\bcc{\begin{center}}                \def\ecc{\end{center}}
\def\bfig{\begin{figure}}               \def\efig{\end{figure}}
\def\bver{\begin{verb}}					\def\ever{\end{verb}}
\def\beq{\begin{equation}}				\def\eeq{\end{equation}}
\def\bmult{\begin{multline}}			\def\emult{\end{multline}}
\def\balign{\begin{align}}				\def\ealign{\end{align}}
\def\beqn{\begin{eqnarray}}             \def\eeqn{\end{eqnarray}}
\def\beqnx{\begin{eqnarray*}}           \def\eeqnx{\end{eqnarray*}}
\def\bbib{\begin{thebibliography}{999}}	\def\ebib{\end{thebibliography}}

\def\bm#1{\mbox{\boldmath $#1$}}

\def\zerob{{\bm 0}}
\def\oneb{{\bm 1}}

\def\ab{{\bm a}}
\def\bb{{\bm b}}
\def\cb{{\bm c}}
\def\db{{\bm d}}
\def\eb{{\bm e}}
\def\fb{{\bm f}}
\def\gb{{\bm g}}
\def\hb{{\bm h}}
\def\ib{{\bm i}}
\def\jb{{\bm j}}
\def\kb{{\bm k}}
\def\lb{{\bm l}}
\def\mb{{\bm m}}
\def\nb{{\bm n}}
\def\ob{{\bm o}}
\def\pb{{\bm p}}
\def\qb{{\bm q}}
\def\rb{{\bm r}}
\def\sb{{\bm s}}
\def\tb{{\bm t}}
\def\ub{{\bm u}}
\def\vb{{\bm v}}
\def\wb{{\bm w}}
\def\xb{{\bm x}}
\def\yb{{\bm y}}
\def\zb{{\bm z}}

\def\Ab{{\bm A}}
\def\Bb{{\bm B}}
\def\Cb{{\bm C}}
\def\Db{{\bm D}}
\def\Eb{{\bm E}}
\def\Fb{{\bm F}}
\def\Gb{{\bm G}}
\def\Hb{{\bm H}}
\def\Ib{{\bm I}}
\def\Jb{{\bm J}}
\def\Kb{{\bm K}}
\def\Lb{{\bm L}}
\def\Mb{{\bm M}}
\def\Nb{{\bm N}}
\def\Ob{{\bm O}}
\def\Pb{{\bm P}}
\def\Qb{{\bm Q}}
\def\Rb{{\bm R}}
\def\Sb{{\bm S}}
\def\Tb{{\bm T}}
\def\Ub{{\bm U}}
\def\Vb{{\bm V}}
\def\Wb{{\bm W}}
\def\Xb{{\bm X}}
\def\Yb{{\bm Y}}
\def\Zb{{\bm Z}}

\def\alphab{\bm{\alpha}}
\def\betab{\bm{\beta}}
\def\deltab{\bm{\delta}}
\def\etab{\bm{\eta}}
\def\epsilonb{\bm{\epsilon}}
\def\gammab{\bm{\gamma}}
\def\omegab{\bm{\omega}}
\def\etab{\bm{\eta}}
\def\thetab{\bm{\theta}}
\def\xib{\bm{\xi}}
\def\lambdab{\bm{\lambda}}
\def\taub{\bm{\tau}}
\def\phib{\bm{\phi}}
\def\mub{\bm{\mu}}
\def\psib{\bm{\psi}}
\def\chib{\bm{\chi}}
\def\sigmab{\bm{\sigma}}

\def\Deltab{\bm{\Delta}}
\def\Lambdab{\bm{\Lambda}}
\def\Phib{\bm{\Phi}}
\def\Psib{\bm{\Psi}}
\def\Sigmab{\bm{\Sigma}}

\def\Ac{{\cal A}}
\def\Bc{{\cal B}}
\def\Cc{{\cal C}}
\def\Dc{{\cal D}}
\def\Ec{{\cal E}}
\def\Fc{{\cal F}}
\def\Gc{{\cal G}}
\def\Hc{{\cal H}}
\def\Ic{{\cal I}}
\def\Jc{{\cal J}}
\def\Kc{{\cal K}}
\def\Lc{{\cal L}}
\def\Mc{{\cal M}}
\def\Nc{{\cal N}}
\def\Oc{{\cal O}}
\def\Pc{{\cal P}}
\def\Qc{{\cal Q}}
\def\Rc{{\cal R}}
\def\Sc{{\cal S}}
\def\Tc{{\cal T}}
\def\Uc{{\cal U}}
\def\Vc{{\cal V}}
\def\Wc{{\cal W}}
\def\Xc{{\cal X}}
\def\Yc{{\cal Y}}
\def\Zc{{\cal Z}}

\def\disp#1{\displaystyle{#1}}

\def\ra{\rightarrow}
\def\la{\leftarrow}
\def\da{\downarrow}
\def\ua{\uparrow}

\def\Ra{\Rightarrow}
\def\La{\Leftarrow}
\def\Da{\Downarrow}
\def\Ua{\Uparrow}

\def\lra{\longrightarrow}
\def\lla{\longleftarrow}
\def\Lra{\Longrightarrow}
\def\Lla{\longleftarrow}

\def\lrarr{\leftrightarrow}
\def\Lrarr{\Leftrightarrow}
\def\udarr{\updownarrow}
\def\Uparr{\Updownarrow}

\def\d#1{\,\mbox{d} #1}
\def\iii{\int_{-\infty}^{+\infty}}
\def\izi{\int_{0}^{\infty}}
\def\izpi{\int_{0}^{\pi}}
\def\izdpi{\int_{0}^{2\pi}}
\def\intd{\int\kern-.8em\int}
\def\intt{\int\kern-.8em\int\kern-.8em\int}
\def\intg{\int\kern-1.1em\int}

\def\wh{\widehat}
\def\xh{\widehat{x}}
\def\thetah{\widehat{\theta}}
\def\betah{\widehat{\beta}}

\def\xbh{\widehat{\xb}}
\def\ybh{\widehat{\yb}}
\def\thetabh{\widehat{\thetab}}
\def\etabh{\widehat{\etab}}
\def\betabh{\widehat{\betab}}

\def\xbhk{\widehat{\xb}^{k}}
\def\thetahk{\widehat{\theta}^{k}}
\def\betahk{\widehat{\beta}^{k}}

\def\xbhkp{\widehat{\xb}^{k+1}}
\def\thetahkp{\widehat{\theta}^{k+1}}
\def\betahkp{\widehat{\beta}^{k+1}}

\def\thetabhk{\widehat{\thetab}^{k}}
\def\betabhk{\widehat{\betab}^{k}}

\def\thetabhkp{\widehat{\thetab}^{k+1}}
\def\betabhkp{\widehat{\betab}^{k+1}}

\def\thetamin{\theta_{\mbox{\tiny min}}}
\def\thetamax{\theta_{\mbox{\tiny max}}}
\def\betamin{\beta_{\mbox{\tiny min}}}
\def\betamax{\beta_{\mbox{\tiny max}}}

\def\expf#1{\exp\left[ {#1} \right]}

\def\argmax#1#2{\arg\max_{#1}\left\{ #2 \right\}}
\def\argmin#1#2{\arg\min_{#1}\left\{ #2 \right\}}

\def\Prob#1{\mbox{Pr}\left\{#1\right\}}
\def\var#1{\mbox{Var}\left\{#1\right\}}
\def\cov#1{\mbox{Cov}\left\{#1\right\}}
\def\corr#1{\mbox{Corr}\left\{#1\right\}}
\def\rang#1{\mbox{rang}\left\{#1\right\}}
\def\det#1{\mbox{d\'et}\left\{#1\right\}}

\def\gbu{\underline{\gb}}
\def\fbu{\underline{\fb}}
\def\zbu{\underline{\zb}}
\def\epsilonbu{\underline{\epsilonb}}
\def\thetabu{\underline{\thetab}}

\def\ve{{v_{\epsilon}}}
\def\sigmae{{\sigma_{\epsilon}}}
\def\Sigmae{{\Sigmab_{\epsilonb}}}
\def\Sigmaf{{\Sigmab_{\fb}}}
\def\Sigmag{{\Sigmab_{\gb}}}

\def\fbuh{\widehat{\fbu}}
\def\zbuh{\widehat{\fbu}}
\def\thetabuh{\widehat{\thetabu}}

\def\fbh{\widehat{\fb}}
\def\Pbh{\widehat{\Pb}}
\def\Sigmabh{\widehat{\Sigmab}}
\def\Abh{\widehat{\Ab}}
\def\Bbh{\widehat{\Bb}}
\def\thetabh{\widehat{\thetab}}
\def\Rbe{\Rb_{\epsilon}}
\def\Rbeh{\widehat{\Rbe}}

\def\Rjk{{R_j}_k}
\def\fbik{{\fb_i}_k}
\def\fbjk{{\fb_j}_k}

\def\mik{{m_i}_k}
\def\sigmaik{{\sigma_i^2}_k}
\def\Sigmaik{{\Sigmab_i}_k}
\def\alphaik{{\alpha_i}_k}
\def\betaik{{\beta_i}_k}

\def\muik{{\mu_i}_k}
\def\vik{{v_i}_k}

\def\mjk{{m_j}_k}
\def\sigmajk{{\sigma_j^2}_k}
\def\Sigmajk{{\Sigmab_j}_k}
\def\alphajk{{\alpha_j}_k}
\def\betajk{{\beta_j}_k}

\def\mujk{{\mu_j}_k}
\def\vjk{{v_j}_k}
\def\njk{{n_j}_k}

\def\alphae{\alpha_{\epsilon}}
\def\betae{\beta_{\epsilon}}
\def\alphaez{\alpha_{\epsilon_0}}
\def\betaez{\beta_{\epsilon_0}}
\def\alphaei{\alpha_{\epsilon_i}}
\def\betaei{\beta_{\epsilon_i}}
\def\alphaeiz{\alpha_{\epsilon_{i_0}}}
\def\betaeiz{\beta_{\epsilon_{i-0}}}

\def\mbz{{{\mb}_z}}
\def\Sigmabz{{{\Sigmab}_z}}
\def\diag#1{\mbox{diag}\left[#1\right]}

\def\sigmah{\widehat{\sigma}}
\def\fh{\widehat{f}}
\def\sigmaz{\sigma_z}

\def\sumi{\sum_{i=1}^{M} }
\def\sumj{\sum_{j=1}^{N} }
\def\lambdah{\wh{\lambda}}
\def\lambdabh{\wh{\lambdab}}

\def\det#1{\hbox{det}\left(#1\right)}
\def\defined{\,\shortstack{$\triangle$\\ =}\,}

\def\zmat{\left[\matrix{0}\right]}
\def\det#1{\hbox{det}(#1)}
\def\th{\widehat{t}}
\def\xh{\widehat{x}}
\def\lambdah{\widehat{\lambda}}
\def\muh{\widehat{\mu}}
\def\sigmah{\widehat{\sigma}}
\def\rhoh{\widehat{\rho}}
\def\betah{\widehat{\beta}}
\def\alphah{\widehat{\alpha}}
\def\tauh{\widehat{\tau}}

\def\tbh{\widehat{\tb}}
\def\mubh{\widehat{\mub}}
\def\sigmabh{\widehat{\sigmab}}
\def\rhobh{\widehat{\rhob}}
\def\oneb{\mbox{\bf 1}}

\def\bm#1{\mbox{\boldmath $#1$}}
\def\zerob{{\bf 0}}
\def\oneb{{\bf 1}}
\def\intg{\int\kern-1.1em\int}
\def\expf#1{\exp\left[ {#1} \right]}

\def\gbu{\underline{\gb}}
\def\fbu{\underline{\fb}}
\def\zbu{\underline{\zb}}
\def\epsilonbu{\underline{\epsilonb}}
\def\uthetab{\underline{\thetab}}

\def\sigmae{{\sigma_{\epsilon}}}
\def\Sigmae{{\Sigma_{\epsilonb}}}
\def\Sigmabe{{\Sigmab_{\epsilonb}}}
\def\Sigmaf{{\Sigmab_{\fb}}}
\def\Sigmag{{\Sigmab_{\gb}}}

\def\fbuh{\widehat{\fbu}}
\def\zbuh{\widehat{\fbu}}
\def\uthetabh{\widehat{\uthetab}}

\def\fbh{\widehat{\fb}}
\def\Sigmabh{\widehat{\Sigmab}}
\def\Abh{\widehat{\Ab}}
\def\thetabh{\widehat{\thetab}}

\def\Rjk{{R_j}_k}
\def\fbik{{\fb_i}_k}
\def\fbjk{{\fb_j}_k}

\def\mik{{m_i}_k}
\def\sigmaik{{\sigma_i^2}_k}
\def\Sigmaik{{\Sigmab_i}_k}
\def\alphaik{{\alpha_i}_k}
\def\betaik{{\beta_i}_k}

\def\muik{{\mu_i}_k}
\def\vik{{v_i}_k}

\def\mjk{{m_j}_k}
\def\sigmajk{{\sigma_j^2}_k}
\def\Sigmajk{{\Sigmab_j}_k}
\def\alphajk{{\alpha_j}_k}
\def\betajk{{\beta_j}_k}

\def\mujk{{\mu_j}_k}
\def\vjk{{v_j}_k}
\def\njk{{n_j}_k}

\def\mbz{{{\mb}_z}}
\def\Sigmabz{{{\Sigmab}_z}}

\def\vect{\mbox{vect}}
\def\lra{\longrightarrow}

\def\gir{g_i(\rb)}
\def\fjr{f_j(\rb)}
\def\zjr{z_j(\rb)}
\def\epsilonir{\epsilon_i(\rb)}
\def\gbr{\gb(\rb)}
\def\fbr{\fb(\rb)}
\def\zbr{\zb(\rb)}
\def\epsilonbr{\epsilonb(\rb)}
\def\fbhr{\fbh(\rb)}
\def\Sigmabhr{\Sigmabh(\rb)}

\def\mbzr{\mb_{z(\rb)}}
\def\Sigmabzr{\Sigmab_{z(\rb)}}
\def\Sigmagbzr{{\Sigmab_g}_{z(\rb)}}

\def\mjzjr#1{{m_{#1}}_{z_{#1}(\rb)}}
\def\sigmajzjr#1{{\sigma_{#1}}_{z_{#1}(\rb)}}

\def\Beta{B}
\def\Betab{\bm{\Beta}}
\def\Betabe{\bm{\Beta}_{\epsilonb}}

\def\fh{\widehat{f}}
\def\sigmah{\widehat{\sigma}}
\def\sigmaei{\sigma_{\epsilon_i}^2}

\def\thetah{\widehat{\theta}}
\def\sigmah{\widehat{\sigma}}

\def\Ftxm{\widetilde{F}_{X}(x_-)}
\def\Ftxp{\widetilde{F}_{X}(x_+)}

\def\NU{{\cal V}}

\def\tFx{\widetilde{F}_{X}}

\def\Fx{F_{X}(x)}
\def\fx{f_{X}(x)}

\def\Fxv{F_{X}(x)}
\def\fxv{f_{X}(x)}

\def\Fxi{{F}_{X_i}(x_i)}
\def\fxi{{f}_{X_i}(x_i)}

\def\Gnu{G_{\NU}(\nu)}
\def\gnu{g_{\NU}(\nu)}

\def\Fnu{F_{\NU}(\nu)}
\def\fnu{f_{\NU}(\nu)}

\def\Finnu#1{F^{-1}_{\NU}(#1)}

\def\Gnua#1{G_{\NU}(#1)}
\def\Gnub#1{G_{\NU}^{-1}(#1)}
\def\Gnuh{G_{\NU}^{-1}(1/2)}

\def\Ftx{\widetilde{F}_{X}(x)}
\def\ftx{\widetilde{f}_{X}(x)}

\def\Ftxi{\widetilde{F}_{X_i}(x_i)}
\def\ftxi{\widetilde{f}_{X_i}(x_i)}

\def\Ftxv{\widetilde{F}_{X}(x)}
\def\FtX#1{\widetilde{F}_{X}(#1)}
\def\ftxv{\widetilde{f}_{X}(x)}

\def\fxnu{f_{X|\NU}(x|\nu)}
\def\fxNU{f_{X|\NU}(x|\NU)}
\def\FxNU{F_{X|\NU}(x|\NU)}

\def\fxNUtheta{f_{X|\NU,\theta}(x|\NU,\theta)}
\def\FxNUtheta{F_{X|\NU,\theta}(x|\NU,\theta)}

\def\Fxnu{F_{X|\NU}(x|\nu)}
\def\Fxnui#1{F_{X|\NU}(x|\nu_#1)}

\def\FxN#1{F_{X|\NU}(#1)}
\def\FxNU{F_{X|\NU}(x|\NU)}
\def\Fxnun#1{F_{X|\NU}(x|#1)}
\def\Fxinun#1{F_{X_i|\NU}(x_i|#1)}
\def\fxnu{f_{X|\NU}(x|\nu)}
\def\fxNU{f_{X|\NU}(x|\NU)}

\def\Fxvnu{F_{X|\NU}(x|\nu)}
\def\Fxvnun#1{F_{X|\NU}(x|#1)}
\def\FxvNU{F_{X|\NU}(x|\NU)}

\def\FXvnu{F_{X|\NU}}

\def\fxvnu{f_{X|\NU}(x|\nu)}
\def\fxvNU{f_{X|\NU}(x|\NU)}

\def\FXvNU#1{{F}_{X|\NU}(#1|\NU)}
\def\FXvn#1{{F}_{X|\NU}(x|#1)}

\def\Fxtheta{F_{X|\theta}(x|\theta)}
\def\fxtheta{f_{X|\theta}(x|\theta)}

\def\Fxvtheta{F_{X|\theta}(x|\theta)}
\def\fxvtheta{f_{X|\theta}(x|\theta)}

\def\Ftxtheta{\widetilde{F}_{X|\theta}(x|\theta)}
\def\ftxtheta{\widetilde{f}_{X|\theta}(x|\theta)}

\def\Ftxvtheta{\widetilde{F}_{X|\theta}(x|\theta)}
\def\ftxvtheta{\widetilde{f}_{X|\theta}(x|\theta)}

\def\Fxnutheta{F_{X|\NU,\theta}(x|\nu,\theta)}
\def\fxnutheta{f_{X|\NU,\theta}(x|\nu,\theta)}

\def\ftheta{f_{\Theta}(\theta)}
\def\fthetaxnuz{f_{\Theta|X,\nu_0}(\theta|x,\nu_0)}
\def\fthetax{f_{\Theta|X}(\theta|x)}

\def\Fxvnutheta{F_{X|\NU,\theta}(x|\nu,\theta)}
\def\FxvNUtheta{F_{X|\NU,\theta}(x|\NU,\theta)}
\def\fxvnutheta{f_{X|\NU,\theta}(x|\nu,\theta)}
\def\Ftxvnutheta{\widetilde{F}_{X|\theta}(x|\theta)}
\def\ftxvnutheta{\widetilde{f}_{X|\theta}(x|\theta)}

\def\FxvNUm{F_{X|\NU}(x_-|\NU)}
\def\FxvNUp{F_{X|\NU}(x_+|\NU)}

\def\Fxinu{F_{X_i|\NU}(x_i|\nu)}
\def\fxinu{f_{X_i|\NU}(x_i|\nu)}

\def\Ftxi{\widetilde{F}_{X_i}(x_i)}
\def\ftxi{\widetilde{f}_{X_i}(x_i)}

\def\fxitheta{f_{X_i|\theta}(x_i|\theta)}
\def\ftxitheta{\widetilde{f}_{X_i|\theta}(x_i|\theta)}

\def\thetah{\widehat{\theta}}
\def\thetat{\widetilde{\theta}}

\def\lnuxv{l_{\nu|X}(\nu|x)}

\def\Lnuxva#1{L_{\nu|X}(#1 | x)}
\def\lnuxva#1{l_{\nu|X}(#1 | x)}

\def\Prob#1{\mbox{P}\left(#1\right)}

\def\lthetaxvnuz{l(\theta)}

\def\lthetaxvnuz{l_{\theta|x,\nu_0}(\theta | x,\nu_0)}
\def\fxnuztheta{f_{X|\nu_0,\theta}(x | \nu_0,\theta)}
\def\fxinuztheta{f_{X|\nu_0,\theta}(x_i | \nu_0,\theta)}
\def\fxvtheta{f_{X|\theta}(x | \theta)}
\def\lthetaxv{l_{\theta|X}(\theta | x)}
\def\ltthetaxv{\widetilde{l}_{\theta|X}(\theta | x)}

\def\ltheta{l(\theta )}
\def\lttheta{\widetilde{l}(\theta)}
\def\Lnu{L(\nu )}
\def\Lnui#1{L(\nu_#1)}
\def\Linu{L_i(\nu )}
\def\Linu{L_i(\nu )}
\def\LNU{L(\NU )}
\def\Lin#1{L^{-1}(#1)}
\def\L#1{L(#1)}
\def\Li#1{L_i(#1)}
\def\FNU#1{F_\NU(#1)}
\def\FinNU#1{F^{-1}_\NU(#1)}

\def\ftthetax{\tilde{f}_{\Theta|X}(\theta|x)}
\def\ER{R}

\def\Ndd#1#2#3{{\cal N}\left( #1 ;~ #2, #3 \right)}
\def\Cdd#1#2#3{{\cal C}\left( #1 ;~ #2, #3 \right)}
\def\Sdd#1#2#3#4{{\cal S}\left( #1 ;~ #2, #3, #4 \right)}
\def\Edd#1#2{{\cal E}\left( #1 ;~ #2 \right)}
\def\DEdd#1#2{{\cal DE}\left( #1 ;~ #2 \right)}
\def\Gdd#1#2#3{{\cal G}\left( #1 ;~ #2, #3 \right)}
\def\IGdd#1#2#3{{\cal IG}\left( #1 ;~ #2, #3 \right)}

\def\Xwr{X(\omega,\rb)}
\def\xwr{x(\omega,\rb)}

\def\muzw{\mu_{z}(\omega)}
\def\muzwr{\mu_{z(\rb)}(\omega)}
\def\szw{\sigma_{z}^2(\omega)}
\def\szwr{\sigma_{z(\rb)}^2(\omega)}
\def\pzw{p_{z}(\omega)}
\def\pzwr{p_{z(\rb)}(\omega)}

\def\hszw{\widehat{\sigma_{z}^2}(\omega)}
\def\hmuzw{\widehat{\mu}_{z}(\omega)}
\def\hpzw{\widehat{p}_{z}(\omega)}

\def\muzwrl{\mu_{z(\rb)}(\omega-l)}

\def\pxwrcz{p\left(\xwr ~|~ z\right)}
\def\pxwr{p\left(\xwr\right)}

\def\Xbw{\Xb(\omega)}
\def\xbw{\xb(\omega)}
\def\uXb{\underline{\Xb}}
\def\uxb{\underline{\xb}}
\def\pxbw{p(\xbw)}
\def\puxb{p(\uxb)}
\def\pzb{P(\zb)}

\def\Szw{S_z(\omega)}
\def\mubw{\mub(\omega)}

\def\Srw{S_{\rb}(\omega)}
\def\Swr{S_{\omega}(\rb)}
\def\Sbw{\Sb(\omega)}
\def\sbw{\sb(\omega)}
\def\Sbr{\Sb(\rb)}
\def\mubw{\mub(\omega)}
\def\cov#1{\mbox{cov}[#1]}

\def\pdf{\emph{pdf}}
\def\pmf{\emph{pmf}}
\def\dpdx#1#2{\frac{\partial #1}{\partial #2}}

\def\REM#1{}

\def\XXwr{X(\omega,\rb)}
\def\xxwr{x(\omega,\rb)}

\def\Xwr{X_{\omega}(\rb)}
\def\xwr{x_{\omega}(\rb)}
\def\Xrw{X_{\rb}(\omega)}
\def\xrw{x_{\rb}(\omega)}

\def\Ywr{Y_{\omega}(\rb)}
\def\ywr{y_{\omega}(\rb)}
\def\Yrw{Y_{\rb}(\omega)}
\def\yrw{y_{\rb}(\omega)}

\def\Ewr{E_{\omega}(\rb)}
\def\ewr{\epsilon_{\omega}(\rb)}
\def\Erw{E_{\rb}(\omega)}
\def\erw{\epsilon_{\rb}(\omega)}

\def\Xbr{\Xb(\rb)}
\def\xbr{\xb(\rb)}
\def\Xbw{\Xb(\omega)}
\def\xbw{\xb(\omega)}

\def\uXb{\underline{\Xb}}
\def\uxb{\underline{\xb}}
\def\uXbz{\underline{\Xb}_z}
\def\uxbz{\underline{\xb}_z}

\def\uthetab{\underline{\thetab}}
\def\uthetabz{\underline{\thetab}_z}

\def\Ywr{Y_{\omega}(\rb)}
\def\ywr{y_{\omega}(\rb)}
\def\Yrw{Y_{\rb}(\omega)}
\def\yrw{y_{\rb}(\omega)}
\def\Ybr{\Yb(\rb)}
\def\ybr{\yb(\rb)}
\def\Ybw{\Yb(\omega)}
\def\ybw{\yb(\omega)}
\def\uYb{\underline{\Yb}}
\def\uyb{\underline{\yb}}
\def\uYbz{\underline{\Yb}_z}
\def\uybz{\underline{\yb}_z}

\def\Ewr{E_{\omega}(\rb)}
\def\ewr{\epsilon_{\omega}(\rb)}
\def\Erw{E_{\rb}(\omega)}
\def\erw{\epsilon_{\rb}(\omega)}
\def\Ebr{Eb(\rb)}
\def\ebr{\epsilonb(\rb)}
\def\Ebw{Eb(\omega)}
\def\ebw{\epsilonb(\omega)}
\def\uEb{\underline{\Eb}}

\def\Erwp{E_{\rb}(\omega')}
\def\Xrwp{X_{\rb}(\omega')}

\def\xwmr{x_{\omega-1}(\rb)}
\def\muzwm{\mu_{z}(\omega-1)}

\def\rhoz{\rho_z}
\def\sez{\sigma_{\eta_z}^2}

\def\xwrp{x_{\omega}(\rb')}
\def\xrwp{x_{\rb}(\omega')}
\def\xwpr{x_{\omega'}(\rb)}
\def\xwprp{x_{\omega'}(\rb')}
\def\Xwrp{x_{\omega}(\rb')}
\def\Xwrp{X_{\omega}(\rb')}
\def\Xwpr{X_{\omega'}(\rb)}
\def\Xwprp{X_{\omega'}(\rb')}

\def\sigmaed{\sigmae^2}
\def\sigmaew{\sigmae(\omega)}
\def\sigmaedw{\sigmaed(\omega)}

\def\Sigmabe{\Sigmab_{\epsilon}}
\def\sigmabh{\widehat{\Sigmab}}
\def\xbh{\widehat{\xb}}
\def\zbh{\widehat{\zb}}
\def\qbh{\widehat{\qb}}
\def\sbh{\widehat{\sb}}

\def\espx#1#2{\mbox{E}_{#1}\left\{#2\right\}}
\def\esp#1{\mbox{E}\left\{#1\right\}}
\def\d#1{\mbox{~d}#1}

\def\Tr#1{\mbox{Tr}\left[#1\right]}

\def\pmatrix#1{\left[\barr{ccccccccccccccc} #1 \earr\right]}
\def\Bf{\bm{B}}
\def\gbh{\widehat{\gb}}
\def\dbh{\widehat{\db}}
\def\Tr#1{\mbox{Tr}\left[#1\right]}
\def\Cov#1{\mbox{Cov}\left[#1\right]}
\def\iid{i.i.d.~}

\def\cro#1{\left[#1\right]}
\def\acc#1{\left\{#1\right\}}
\def\aprio{\emph{a priori~}}
\def\apost{\emph{a posteriori~}}
\def\hbh{\widehat{\hb}}
\def\qh{\widehat{q}}
\def\fbt{\widetilde{\fb}}
\def\thetabt{\widetilde{\thetab}}

\def\mubt{\widetilde{\mub}}
\def\Sigmabt{\widetilde{\Sigmab}}
\def\qt{\widetilde{q}}
\def\vt{\widetilde{v}}
\def\zt{\widetilde{z}}
\def\Dbt{\widetilde{\Db}}
\def\alphat{\widetilde{\alpha}}
\def\betat{\widetilde{\beta}}
\def\abt{\widetilde{\ab}}
\def\ajt{\widetilde{a}_j}
\def\taut{\widetilde{\tau}}
\def\mut{\widetilde{\mu}}
\def\Zbt{\widetilde{\Zb}}
\def\Abt{\widetilde{\Ab}}
\def\zbt{\widetilde{\zb}}

\def\bloca#1#2#3{
\begin{picture}(50,50)
  \put(32,42){\makebox(0,0){#1}}
  \put(25,50){\vector(0,-1){20}}
  \put(0,10){\framebox(50,20){#2}}
  \put(0,0){\makebox(50,0){#3}}  
\end{picture}
}

\def\ugb{\underline{\gb}}
\def\fbh{\widehat{\fb}}
\def\zbh{\widehat{\zb}}
\def\qbh{\widehat{\qb}}
\def\thetabh{\widehat{\thetab}}
\def\Pbh{\widehat{\Pb}}

\def\sigmae{\sigma_{\epsilon}}
\def\REM#1{}
\def\fh{\widehat{f}}
\def\zh{\widehat{z}}
\def\disp{\displaystyle}
\def\oneb{{\boldmath{1}}}
\def\zerob{{\boldmath{0}}}
\def\d#1{\; \mbox{d} #1}
\def\expf#1{\exp\left\{#1\right\}}
\def\esp#1{\mbox{E}\left\{#1\right\}}
\def\lra{\longrightarrow}

\def\bslide#1{\begin{frame}{\Large\bf #1}}
\def\eslide{\end{frame}}

\def\inversions#1#2#3{
\begin{picture}(100,40)
  \put(0,15){\makebox(15,0){#1}}
  \put(20,15){\vector(1,0){10}}
  \put(30,7){\framebox(40,15){#2}}
  \put(70,15){\vector(1,0){10}}
  \put(85,15){\makebox(15,0){#3}}
\end{picture}
}

\def\blue#1{{\color{blue}#1}}
\def\red#1{{\color{red}#1}}
\def\green#1{{\color{green}#1}}

\def\rfb{\red{\fb}}
\def\bgb{\blue{\gb}}
\def\rfbh{\red{\widehat{\fb}}}
\def\rfbt{\red{\widetilde{\fb}}}
\def\rzbt{\red{\widetilde{\zb}}}
\def\rqb{\red{\qb}}
\def\rqbh{\red{\widehat{\qb}}}
\def\rzb{\red{\zb}}
\def\rzbh{\red{\widehat{\zb}}}
\def\rthetab{\red{\thetab}}
\def\retab{\red{\etab}}
\def\rthetabh{\red{\widehat{\thetab}}}
\def\rthetabt{\red{\widetilde{\thetab}}}
\def\retabt{\red{\widetilde{\etab}}}
\def\ralphabt{\red{\widetilde{\alphab}}}

\def\ugb{\underline{\gb}}
\def\fbh{\widehat{\fb}}
\def\zbh{\widehat{\zb}}
\def\qbh{\widehat{\qb}}
\def\thetabh{\widehat{\thetab}}
\def\Pbh{\widehat{\Pb}}

\def\UP#1{\uppercase{#1}}
\def\sca#1{\uppercase{#1}}

\def\sigmae{\sigma_{\epsilon}}
\def\REM#1{}
\def\fh{\widehat{f}}
\def\zh{\widehat{z}}
\def\disp{\displaystyle}
\def\oneb{{\boldmath{1}}}
\def\zerob{{\boldmath{0}}}
\def\d#1{\; \mbox{d} #1}
\def\expf#1{\exp\left\{#1\right\}}
\def\esp#1{\mbox{E}\left\{#1\right\}}
\def\lra{\longrightarrow}

\def\rfb{\red{\fb}}
\def\rFb{\red{\Fb}}
\def\bgb{\blue{\gb}}
\def\rfbh{\red{\widehat{\fb}}}
\def\rqb{\red{\qb}}
\def\rqbh{\red{\widehat{\qb}}}
\def\rzb{\red{\zb}}
\def\rzbh{\red{\widehat{\zb}}}
\def\rthetab{\red{\thetab}}
\def\ralphab{\red{\alphab}}
\def\rbetab{\red{\betab}}
\def\rthetabh{\red{\widehat{\thetab}}}
\def\ralphabh{\red{\widehat{\alphaab}}}
\def\rSigmab{\red{\Sigmab}}
\def\rSigmabe{\red{\Sigmabe}}

\def\rz{\red{z}}
\def\rzh{\red{\widehat{z}}}

\def\rab{\red{\ab}}
\def\rabh{\red{\widehat{\ab}}}
\def\rAb{\red{\Ab}}
\def\rAbh{\red{\widehat{\Ab}}}
\def\rBb{\red{\Bb}}
\def\rBbh{\red{\widehat{\Bb}}}
\def\rPbh{\red{\widehat{\Pb}}}
\def\rub{\red{\ub}}
\def\rubh{\red{\widehat{\ub}}}

\def\rVb{\red{\Vb}}
\def\rVbh{\widehat{\rVb}}
\def\rVbt{\widetilde{\rVb}}

\def\rvb{\red{\vb}}
\def\rvbh{\widehat{\rvb}}
\def\rvbt{\widetilde{\rvb}}

\def\bfb{\blue{\fb}}
\def\bFb{\blue{\Fb}}

\def\rfi{\red{f}_i}
\def\rfj{\red{f}_j}
\def\ruj{\red{u}_j}
\def\raj{\red{a}_j}
\def\rzj{\red{z}_j}
\def\rvj{\red{v}_j}
\def\rtau{\red{\tau}}
\def\rtauj{\red{\tau_j}}
\def\rtaub{\red{\taub}}
\def\rtauh{\widehat{\rtau}}
\def\rtaujh{\widehat{\rtauj}}
\def\rtaubh{\widehat{\rtaub}}
\def\rve{\red{v_{\epsilon}}}
\def\rv{\red{v}}

\def\rfjh{\red{\widehat{f}}_j}
\def\rujh{\red{\widehat{u}}_j}
\def\rajh{\red{\widehat{a}}_j}
\def\rzjh{\red{\widehat{z}}_j}
\def\rvjh{\red{\widehat{v}}_j}

\def\ralpha{\red{\alpha}}
\def\rbeta{\red{\beta}}
\def\rtheta{\red{\theta}}

\def\rlambda{\red{\lambda}}
\def\ralphah{\widehat{\ralpha}}
\def\rbetah{\widehat{\rbeta}}
\def\rthetah{\widehat{\rtheta}}
\def\rlambdah{\widehat{\rlambda}}

\def\ubh{\widehat{\ub}}

\def\rve{\red{v_{\epsilon}}}
\def\alphae{\alpha_{\epsilon_0}}
\def\betae{\beta_{\epsilon_0}}
\def\alphajh{\widehat{\alpha}_j}
\def\betajh{\widehat{\beta}_j}
\def\alphaeh{\widehat{\alpha}_{\epsilon_0}}
\def\betaeh{\widehat{\beta}_{\epsilon_0}}

\def\rqb{\red{\qb}}
\def\rqbh{\red{\widehat{\qb}}}
\def\rzb{\red{\zb}}
\def\rZb{\red{\Zb}}
\def\rab{\red{\ab}}
\def\rAb{\red{\Ab}}
\def\rvb{\red{\vb}}
\def\rmb{\red{\mb}}
\def\rzbh{\red{\widehat{\zb}}}
\def\rvbh{\red{\widehat{\vb}}}
\def\rmbh{\red{\widehat{\mb}}}
\def\rthetab{\red{\thetab}}
\def\rthetabh{\red{\widehat{\thetab}}}
\def\rfbh{\widehat{\rfb}}
\def\rhb{\red{\hb}}
\def\rhbh{\widehat{\rhb}}
\def\rthetai{\red{\theta}_i}
\def\rthetab{\red{\thetab}}
\def\rfjm{\red{f}_{j-1}}
\def\rfim{\red{f}_{i-1}}

\def\muh{\widehat{\mu}}
\def\vh{\widehat{v}}

\def\rSigmabh{\red{\Sigmabh}}
\def\rVbh{\red{\Vbh}}
\def\Fbh{\red{\Fb}}
\def\rFbh{\red{\Fbh}}

\def\Sigmabeh{\widehat{\Sigmabe}}
\def\rSigmabeh{\widehat{\red{\Sigmabe}}}

\def\rf{\red{f}}
\def\bg{\blue{g}}
\def\rF{\red{F}}
\def\bG{\blue{G}}

\def\sumk{\sum_{k=1}^K}
\def\suml{\sum_{l=1}^L}
\def\sumn{\sum_{n=1}^N}
\def\sumj{\sum_{j=1}^J}
\def\sumi{\sum_{i=1}^I}

\def\nub{\bm{\nu}}
\def\rak{\red{a_k}}
\def\rbk{\red{b_k}}
\def\rck{\red{c_k}}
\def\rtk{\red{t_k}}
\def\rvk{\red{v_k}}
\def\rnuk{\red{\nu_k}}
\def\rnuz{\red{\nu_0}}
\def\rphik{\red{\phi_k}}
\def\rnub{\red{\nub}}
\def\rtb{\red{\tb}}
\def\rphib{\red{\phib}}
\def\rfn{\red{f_n}}
\def\rtn{\red{t_n}}
\def\rvn{\red{v_n}}

\def\rve{\red{v_\epsilon}}
\def\rvf{\red{v_f}}
\def\rveh{\red{\widehat{v}_\epsilon}}
\def\rvfh{\red{\widehat{v}_f}}

\def\rnu{\red{\nu}}
\def\rphi{\red{\phi}}
\def\rbl{\red{b_l}}
\def\rbb{\red{\bb}}

\def\modela{\tt\setlength{\unitlength}{0.92pt}
\begin{picture}(250,106)
\thinlines    \put(-5,105){$\Nc(t|\red{t_1,v_1})$}
              \put(35,100){\vector(1,0){25}}
              \put(60,95){\framebox(37,16){\red{$a_1$}}}
			  \multiput(77,72)(0,7){3}{\circle*{4}}
              \put(-5,65){$\Nc(t|\red{t_k,v_k})$}
              \put(35,59){\vector(1,0){25}}
              \put(60,51){\framebox(37,16){\red{$a_k$}}}
              \multiput(77,32)(0,7){3}{\circle*{4}}              
              \put(-5,23){$\Nc(t|\red{t_K,v_K})$}
              \put(60,10){\framebox(37,16){\red{$a_K$}}}
              \put(35,18){\vector(1,0){25}}

              \put(98,103){\vector(1,-1){38}}
              \put(98,59){\vector(1,0){39}}
              \put(98,17){\vector(1,1){38}}

              \put(150,60){\circle{24}}
              \put(146,57){$\Large+$}
              \put(165,75){$g_{\rthetab}(t)$}
              \put(164,60){\vector(1,0){33}}
              
              \put(212,85){$\epsilon(t)$}
              \put(210,95){\vector(0,-1){20}}
              \put(211,60){\circle{24}}
              \put(207,58){$\Large+$}
              
              \put(225,66){$g(t)$}
              \put(225,60){\vector(1,0){20}}
\end{picture}
}

\def\modelb{\tt\setlength{\unitlength}{0.92pt}
\begin{picture}(250,106)
\thinlines    \put(-5,105){$\delta(t-\red{t_1})$}
              \put(0,100){\vector(1,0){37}}
              \put(38,95){\framebox(60,16){$\red{a_1} \Nc(t|0,\red{v_1})$}}
			  \multiput(77,72)(0,7){3}{\circle*{4}}
              \put(-5,65){$\delta(t-\red{t_k})$}
              \put(0,59){\vector(1,0){37}}
              \put(38,51){\framebox(60,16){$\red{a_k} \Nc(t|0,\red{v_k})$}}
              \multiput(77,32)(0,7){3}{\circle*{4}}              
              \put(-5,23){$\delta(t-\red{t_K})$}
              \put(38,10){\framebox(60,16){$\red{a_K} \Nc(t|0,\red{v_K})$}}
              \put(0,18){\vector(1,0){37}}

              \put(98,103){\vector(1,-1){38}}
              \put(98,59){\vector(1,0){39}}
              \put(98,17){\vector(1,1){38}}

              \put(150,60){\circle{24}}
              \put(146,57){$\Large+$}
              \put(165,75){$g_{\rthetab}(t)$}
              \put(164,60){\vector(1,0){33}}
              
              \put(212,85){$\epsilon(t)$}
              \put(210,95){\vector(0,-1){20}}
              \put(211,60){\circle{24}}
              \put(207,58){$\Large+$}
              
              \put(225,66){$g(t)$}
              \put(225,60){\vector(1,0){20}}
\end{picture}
}

\def\modelc{\tt\setlength{\unitlength}{0.92pt}
\begin{picture}(250,106)
\thinlines    \put(-5,105){$\delta(t-\red{t_1})$}
              \put(0,100){\vector(1,0){37}}
              \put(38,95){\framebox(40,16){$\red{a_1}$}}
			  \multiput(57,72)(0,7){3}{\circle*{4}}
              \put(-5,65){$\delta(t-\red{t_k})$}
              \put(0,59){\vector(1,0){37}}
              \put(38,51){\framebox(40,16){$\red{a_k}$}}
              \multiput(57,32)(0,7){3}{\circle*{4}}              
              \put(-5,23){$\delta(t-\red{t_K})$}
              \put(38,10){\framebox(40,16){$\red{a_K}$}}
              \put(0,18){\vector(1,0){37}}

              \put(78,103){\vector(1,-1){38}}
              \put(78,59){\vector(1,0){39}}
              \put(78,17){\vector(1,1){38}}

              \put(130,60){\circle{24}}
              \put(126,57){$\Large+$}
              \put(144,60){\vector(1,0){10}}
              \put(154,52){\framebox(50,16){$\Nc(t|0,v)$}}
              
              \put(205,75){$g_{\rthetab}(t)$}
              \put(204,60){\vector(1,0){33}}
              
              \put(252,85){$\epsilon(t)$}
              \put(250,95){\vector(0,-1){20}}
              \put(251,60){\circle{24}}
              \put(247,58){$\Large+$}
              
              \put(265,66){$g(t)$}
              \put(265,60){\vector(1,0){20}}
\end{picture}
}

\def\modeld{\tt\setlength{\unitlength}{0.92pt}
\begin{picture}(250,106)
\thinlines    \put(20,70){$s(t)=\sum_k \rak \delta(t-\red{t_k})$}
              \put(105,60){\vector(1,0){50}}
              \put(154,52){\framebox(50,16){$\Nc(t|0,v)$}}
              
              \put(205,75){$g_{\rthetab}(t)$}
              \put(204,60){\vector(1,0){33}}
              
              \put(252,85){$\epsilon(t)$}
              \put(250,95){\vector(0,-1){20}}
              \put(251,60){\circle{24}}
              \put(247,58){$\Large+$}
              
              \put(265,66){$g(t)$}
              \put(265,60){\vector(1,0){20}}
              \put(0,20){\siga}
\end{picture}
}

\def\siga{\setlength{\unitlength}{0.46pt}
\begin{picture}(290,108)
\thinlines    \put(10,26){\vector(1,0){260}}
              \put(16,12){\vector(0,1){88}}
              \put(48,27){\vector(0,1){34}}
              \put(79,27){\vector(0,1){54}}
              \put(130,27){\vector(0,1){46}}
              \put(188,26){\vector(0,1){40}}
              \put(262,14){$t$}
              \put(182,16){$\red{t_K}$}
              \put(195,63){$\red{a_K}$}
              \put(125,15){$\red{t_k}$}
              \put(137,68){$\red{a_k}$}
              \put(75,14){$\red{t_2}$}
              \put(84,73){$\red{a_2}$}
              \put(43,15){$\red{t_1}$}
              \put(54,56){$\red{a_1}$}
              \multiput(150,46)(8,0){3}{\circle*{4}}
\end{picture}
}

\def\modele{\tt\setlength{\unitlength}{0.92pt}
\begin{picture}(250,106)
\thinlines    \put(20,70){$s(t)=\sum_n \rfn \delta(t-n\Delta t)$}
              \put(105,60){\vector(1,0){50}}
              \put(154,52){\framebox(50,16){$\Nc(t|0,v)$}}
              
              \put(205,75){$g_{\rfb}(t)$}
              \put(204,60){\vector(1,0){33}}
              
              \put(252,85){$\epsilon(t)$}
              \put(250,95){\vector(0,-1){20}}
              \put(251,60){\circle{24}}
              \put(247,58){$\Large+$}
              
              \put(265,66){$g(t)$}
              \put(265,60){\vector(1,0){20}}
              \put(0,20){\sigb}
\end{picture}
}

\def\sigb{\setlength{\unitlength}{0.46pt}
\begin{picture}(290,108)
\thinlines    \put(10,26){\vector(1,0){260}}
              \put(10,12){\vector(0,1){88}}
              
              \put(20,27){\vector(0,1){34}}
              \put(40,27){\vector(0,1){14}}
              \put(60,27){\vector(0,1){54}}
              \put(80,27){\vector(0,1){31}}
              \put(100,27){\vector(0,1){46}}
              \put(120,27){\vector(0,1){40}}
              \put(140,27){\vector(0,1){48}}
              \put(160,27){\vector(0,1){34}}
              \put(180,27){\vector(0,1){14}}
              \put(200,27){\vector(0,1){31}}
              \put(220,27){\vector(0,1){46}}
              \put(240,27){\vector(0,1){40}}
              
              \put(270,14){$t$}
              \put(15,10){$1$}
              \put(15,63){$\red{f_1}$}
              \put(135,10){$n$}
              \put(135,68){$\red{f_n}$}
              \put(235,10){$N$}
              \put(235,68){$\red{f_N}$}

\end{picture}
}

\def\sigbr{\setlength{\unitlength}{0.46pt}
\begin{picture}(290,108)
\thinlines    \put(10,26){\vector(1,0){260}}
              \put(10,12){\vector(0,1){88}}
              
              \put(20,27){\vector(0,1){34}}
              \put(40,27){\vector(0,1){14}}
              \put(60,27){\vector(0,1){54}}
              \put(80,27){\vector(0,1){31}}
              \put(100,27){\vector(0,1){46}}
              \put(120,27){\vector(0,1){40}}
              \put(140,27){\vector(0,1){48}}
              \put(160,27){\vector(0,1){34}}
              \put(180,27){\vector(0,1){14}}
              \put(200,27){\vector(0,1){31}}
              \put(220,27){\vector(0,1){46}}
              \put(240,27){\vector(0,1){40}}
              \put(270,14){$t$}
              \put(15,10){$1$}
              \put(15,63){$\red{f_1}$}
              \put(135,10){$n$}
              \put(135,68){$\red{f_n}$}
              \put(235,10){$N$}
              \put(235,68){$\red{f_N}$}

              \put(60,10){$\red{\Delta t}$}
\end{picture}
}

\def\modelf{\tt\setlength{\unitlength}{0.92pt}
\begin{picture}(250,106)
\thinlines    \put(20,70){$s(t)=\sum_n \rfn \delta(t-n\red{\Delta t})$}
              \put(105,60){\vector(1,0){50}}
              \put(154,52){\framebox(50,16){$\Nc(t|0,\red{v})$}}
              
              \put(205,75){$g_{\rfb,\rthetab}(t)$}
              \put(204,60){\vector(1,0){33}}
              
              \put(252,85){$\epsilon(t)$}
              \put(250,95){\vector(0,-1){20}}
              \put(251,60){\circle{24}}
              \put(247,58){$\Large+$}
              
              \put(265,66){$g(t)$}
              \put(265,60){\vector(1,0){20}}
              \put(0,20){\sigbr}
\end{picture}
}

\def\sig{\tt\setlength{\unitlength}{1.38pt}
\begin{picture}(290,108)
\thinlines    
              \put(10,26){\vector(1,0){260}}
              \put(48,27){\vector(0,1){34}}
              \put(16,12){\vector(0,1){88}}
              \put(79,27){\vector(0,1){54}}
              \put(102,27){\vector(0,1){31}}
              \put(130,27){\vector(0,1){46}}
              \put(188,26){\vector(0,1){40}}
              \put(262,14){$t$}
              \put(22,90){$g(t)$}
              \put(182,16){$t_K$}
              \put(195,63){$a_K$}
              \put(125,15){$t_k$}
              \put(137,68){$a_k$}
              \put(75,14){$t_2$}
              \put(84,73){$a_2$}
              \put(43,15){$t_1$}
              \put(54,56){$a_1$}
              \multiput(150,46)(8,0){3}{\circle*{4}}
\end{picture}
}

\def\decgb{\tt\setlength{\unitlength}{0.92pt}
\begin{picture}(363,122)
\thinlines    \put(248,66){\vector(1,0){14}}
              \put(215,57){\framebox(33,18){$h(t)$}}
              \put(288,66){\vector(1,0){22}}
              \put(270,64){$\large\Sigma$}
              \put(275,67){\circle{26}}
              \put(275,102){\vector(0,-1){20}}
              \put(280,94){$b(t)$}
              \put(296,75){$y(t)$}
              \multiput(100,35)(0,8){3}{\circle*{4}}
              \put(194,72){$x(t)$}
              \put(193,66){\vector(1,0){23}}
              \put(176,61){$\large\Sigma$}
              \put(181,65){\circle{26}}
              \put(80,92){\framebox(43,18){$a_1$}}
              \put(80,58){\framebox(43,18){$a_2$}}
              \put(79,10){\framebox(43,18){$a_K$}}
              \put(123,66){\vector(1,0){45}}
              \put(123,101){\vector(4,-3){44}}
              \put(122,18){\vector(1,1){44}}
              \put(55,101){\vector(1,0){23}}
              \put(55,66){\vector(1,0){23}}
              \put(55,19){\vector(1,0){23}}
              \put(10,104){$g_1(t-t_1)$}
              \put(10,69){$g_2(t-t_2)$}
              \put(10,22){$g_K(t-t_k)$}
\end{picture}
}

\def\decgc{\tt\setlength{\unitlength}{0.92pt}
\begin{picture}(287,68)
\thinlines    \put(36,20){\vector(1,0){23}}
              \put(101,20){\vector(1,0){23}}
              \put(167,20){\vector(1,0){16}}
              \put(124,12){\framebox(43,18){$h(t)$}}
              \put(210,20){\vector(1,0){23}}
              \put(192,20){$\large\Sigma$}
              \put(196,23){\circle{26}}
              \put(196,57){\vector(0,-1){20}}
              \put(200,49){$b(t)$}
              \put(218,30){$y(t)$}
              \put(15,20){$s(t)$}
              \put(101,32){$x(t)$}
              \put(59,11){\framebox(43,18){$g(t)$}}
\end{picture}
}

\def\sigd{
\begin{picture}(290,108)
\thinlines    \put(10,26){\vector(1,0){260}}
              \put(48,27){\vector(0,1){34}}
              \put(16,12){\vector(0,1){88}}
              \put(79,27){\vector(0,1){54}}
              \put(102,27){\vector(0,1){31}}
              \put(130,27){\vector(0,1){46}}
              \put(188,26){\vector(0,1){40}}
              \put(262,14){$t$}
              \put(22,90){$g(t)$}
              \put(182,16){$t_K$}
              \put(195,63){$a_K$}
              \put(125,15){$t_k$}
              \put(137,68){$a_k$}
              \put(75,14){$t_2$}
              \put(84,73){$a_2$}
              \put(43,15){$t_1$}
              \put(54,56){$a_1$}
              \multiput(150,46)(8,0){3}{\circle*{4}}
\end{picture}
}

\def\wt#1{\widetilde{#1}}
\def\Thetab{\bm{\Theta}}
\def\Thetabh{\widehat{\Thetab}}
\def\abh{\widehat{\ab}}
\def\kbt{\tilde{\kb}}
\def\kt{\tilde{k}}
\def\gammat{\tilde{\gamma}}
\def\Sigmat{\tilde{\Sigma}}
\def\etat{\tilde{\eta}}

\def\rHb{\red{\Hb}}
\def\rFb{\red{\Fb}}

\def\wt#1{\widetilde{#1}}
\def\Thetab{\bm{\Theta}}
\def\Thetabh{\widehat{\Thetab}}
\def\abh{\widehat{\ab}}
\def\kbt{\tilde{\kb}}
\def\kt{\tilde{k}}
\def\gammat{\tilde{\gamma}}
\def\Sigmat{\tilde{\Sigma}}
\def\etat{\tilde{\eta}}
\def\zetat{\tilde{\zeta}}
\def\Mt{\tilde{M}}
\def\Mh{\hat{M}}
\def\Nt{\tilde{N}}
\def\atk{{\tilde{a}_k}}
\def\ztk{{\tilde{z}_k}}

\def\rh{\red{h}}
\def\rvbe{\rvb_{\epsilon}}
\def\rvei{\rv_{\epsilon_i}}
\def\rVbe{\rVb_{\epsilon}}
\def\rVbeh{\red{\Vbh}_{\epsilon}}
\def\vbh{\hat{\vb}}
\def\rVbfh{\red{\Vbh}_f}
\def\rvbeh{\rvbh_{\epsilon}}
\def\rvbet{\rvbt_{\epsilon}}
\def\rvbfh{\rvbh_f}

\def\rvbf{\rvb_f}
\def\rvfj{\rv_{f_j}}
\def\rVbf{\red{\Vb}_f}

\def\alphaez{\alpha_{\epsilon_0}}
\def\betaez{\beta_{\epsilon_0}}
\def\alphafz{\alpha_{f_0}}
\def\betafz{\beta_{f_0}}

\def\wh#1{\widehat{#1}}
\def\disp#1{{\displaystyle #1}}

\def\ER{\mbox{I\kern-.25em R}}
\def\EC{\mbox{C\kern-.8em C}}
\def\EZ{\mbox{Z\kern-.55em Z}}
\def\EN{\mbox{N\kern-.8em N}}

\def\bm#1{\mbox{\boldmath $#1$}}
\def\sb{{\bm s}}
\def\ub{{\bm u}}
\def\xb{{\bm x}}
\def\yb{{\bm y}}
\def\db{{\bm{d}}}
\def\Ab{{\bm A}}
\def\Bb{{\bm B}}
\def\Fb{{\bm F}}
\def\Rb{{\bm R}}
\def\Sb{{\bm S}}
\def\Ub{{\bm U}}

\def\thetab{\bm{\theta}}
\def\phib{\bm{\phi}}
\def\lambdab{\bm{\lambda}}
\def\Lambdab{\bm{\Lambda}}
\def\Sigmab{\bm{\Sigma}}
\def\lra{\longrightarrow}
\def\intg{\int\kern-1.1em\int}
\def\intd{\int\kern-.8em\int}

\def\apriori{{\em a priori} }
\def\aposteriori{{\em a posteriori} }

\def\d#1{\,\mbox{d}#1}
\def\esp#1{\mbox{E}\left\{ #1 \right\}}
\def\espx#1#2{\mbox{E}_{#1}\left\{ #2 \right\}}
\def\expf#1{\exp\left[ {#1} \right]}
\def\dpdx#1#2{\frac{\partial #1}{\partial #2}}
\def\dpdxd#1#2{\frac{\partial^2 #1}{\partial #2^2}}

\def\ent#1{\hbox{H}\left[#1\right]}
\def\dent#1{\delta\hbox{H}\left[#1\right]}
\def\rent#1{\hbox{D}\left[#1\right]}
\def\kullback#1{\hbox{K}\left[#1\right]}
\def\pent#1{\hbox{PE}\left[#1\right]}
\def\mdiv#1{\hbox{J}\left[#1\right]}
\def\infmut#1{\hbox{I}\left[#1\right]}
\def\idisc#1{\hbox{ID}\left[#1\right]}
\def\j{\mbox{j}}

\def\pxtheta{p(x | \thetab)}
\def\pxthetas{p(x | \thetab^*)}
\def\pxthetasd{p(x | \thetab^*+\Delta\thetab)}

\def\tr#1{\hbox{tr}\left(#1\right)}
\def\det#1{\hbox{det}\left(#1\right)}

\def\ieeeSSC{IEEE Transactions on Systems Science and Cybernetics}
\def\ieeeSP{IEEE Transactions on Signal Processing}			
\def\ieeeP{Proceedings of the IEEE}					
\def\ieeeIT{IEEE Transactions on Information Theory}			
\def\TS{Traitement du Signal}						
\def\siamCO{SIAM Journal of Control}					
\def\AISM{Annals of Institute of Statistical Mathematics}		
\def\ieeeASSP{IEEE Transactions on Acoustics Speech and Signal Processing}
\def\ieeeAC{IEEE Transactions on Automatic and Control}			

\def\jan{janvier\xspace}
\def\feb{f\'evrier\xspace}
\def\mar{mars\xspace}
\def\apr{avril\xspace}
\def\may{mai\xspace}
\def\jun{juin\xspace}
\def\jul{juillet\xspace}
\def\aug{ao\^ut\xspace}
\def\oct{octobre\xspace}
\def\nov{novembre\xspace}
\def\dec{d\'ecembre\xspace}
\def\Jan{January\xspace}	
\def\Feb{February\xspace}
\def\Mar{March\xspace}
\def\Apr{April\xspace}
\def\May{May\xspace}
\def\Jun{June\xspace}
\def\Jul{July\xspace}
\def\Aug{August\xspace}
\def\Sep{September\xspace}
\def\Oct{October\xspace}
\def\Nov{November\xspace}
\def\Dec{December\xspace}

\def\espq#1{\left<#1\right>_q}

\def\rZbe{\rZb_{\epsilon}}
\def\nubt{\widetilde{\nub}}

\def\rVbf{\rVb_{f}}
\def\rvbf{\rvb_{f}}
\def\rvfj{\red{v}_{f_j}}
\def\rvbhe{\rvbh_{\epsilon}}
\def\rvbhf{\rvbh_{f}}
\def\bsplit{\begin{split}}
\def\esplit{\end{split}}
\def\alphaei{\alpha_{\epsilon_i}}
\def\alphafj{\alpha_{f_j}}
\def\betaei{\beta_{\epsilon_i}}
\def\betafj{\beta_{f_j}}

\def\bgbh{\widehat{\bgb}}
\def\rfh{\widehat{\rf}}
\def\rFh{\widehat{\rF}}
\def\rc{\red{r}}
\def\rch{\widehat{\rc}}
\def\partialx{\frac{\partial }{\partial x}}
\def\partialy{\frac{\partial }{\partial y}}
\def\partialr{\frac{\partial }{\partial r}}
\def\partialxd{\frac{\partial^2 }{\partial x^2}}
\def\partialyd{\frac{\partial^2 }{\partial y^2}}
\def\partialrd{\frac{\partial^2 }{\partial r^2}}
\def\cosphi{\cos(\phi)}
\def\sinphi{\sin(\phi)}
\def\sumk{\sum_{k=1}^K}
\def\sumj{\sum_{j=1}^N}
\def\sumi{\sum_{i=1}^M}
\def\summ{\sum_{m=1}^M}
\def\sumn{\sum_{n=1}^N}
\def\summn{\sum_{m=1}^M\sum_{n=1}^N}

\def\df{\dot{f}}
\def\ddf{\ddot{f}}

\def\rf{\red{f}}
\def\rdf{\red{\df}}
\def\rddf{\red{\ddf}}

\def\rfb{\red{\fb}}
\def\rdfb{\red{\dot{\fb}}}
\def\rddfb{\red{\ddot{\fb}}}

\def\bdg{\blue{\dot{g}}}
\def\bddg{\blue{\ddot{g}}}

\def\bgb{\blue{\gb}}
\def\bdg{\blue{\dot{g}}}

\def\bdgb{\blue{\dot{\gb}}}
\def\bddgb{\blue{\ddot{\gb}}}

\def\epsilonbd{\dot{\epsilonb}}
\def\epsilonbdd{\ddot{\epsilonb}}

\def\rdfbh{\widehat{\rdfb}}
\def\rddfbh{\widehat{\rddfb}}

\def\rqh{\red{\widehat{q}}}
\def\rdfh{\red{\widehat{\dot{f}}}}
\def\rddfh{\red{\widehat{\ddot{f}}}}

\def\vek{v_{\epsilon_k}}
\def\rveih{\widehat{\rve}_i}
\def\rvfjh{\widehat{\rvf}_j}

\def\alphaxiz{{\alpha_{\xi_0}}}
\def\betaxiz{{\beta_{\xi_0}}}
\def\alphazz{{\alpha_{z_0}}}

\def\betazz{{\beta_{z_0}}}
\def\rVbz{{\rVb_z}}
\def\rvbz{{\rvb_z}}
\def\rvxi{{\rv_\xi}}
\def\rvzj{{\rv_{z_j}}}
\def\rvxih{{\widehat{\rv}_\xi}}
\def\rvbzh{{\widehat{\rvb}_z}}
\def\rvzjh{{\widehat{\rv}_{z_j}}}

\def\alphae{{\alpha_\epsilon}}

\def\alphaet{{\widetilde{\alpha}_\epsilon}}
\def\alphaft{{\widetilde{\alpha}_f}}
\def\betae{{\beta_\epsilon}}
\def\betaet{{\widetilde{\beta}_\epsilon}}
\def\betaft{{\widetilde{\beta}_f}}
\def\rfih{\hat{\rf}_i}
\def\rfbk{\rfb^{(k)}}
\def\rfbkp{\rfb^{(k+1)}}

\def\rmh{\hat{\red{m}}}

\def\rmk{{\red{m}_k}}
\def\rmkh{{\hat{\red{m}}_k}}
\def\rvk{{\red{v}_k}}
\def\rvkh{{\hat{\red{v}}_k}}
\def\rak{{\red{a}_k}}
\def\rakh{{\hat{\red{a}}_k}}

\def\rvh{\hat{\red{v}}}
\def\rveh{\hat{\rve}}

\def\rmt{\tilde{\red{m}}}
\def\rvt{\tilde{\red{v}}}
\def\rvet{\tilde{\rve}}
\def\rveih{\red{\hat{v}}_{\epsilon_i}}

\def\rveb{\red{\vb_{\epsilon}}}
\def\rvebh{\red{\vbh_{\epsilon}}}

\def\blocgi{\begin{picture}(80,50)
  \put(60,40){\makebox(0,0){$\red{m}$}}
  \put(60,40){\circle{20}}
  \put(60,30){\vector(0,-1){20}}
  \put(60,0){\circle{20}}
  \put(60,0){\makebox(0,0){$\blue{g}_i$}}
  \put(20,0){\circle{20}}
  \put(20,0){\makebox(0,0){$\epsilon_i$}}
  \put(30,0){\vector(1,0){20}}
\end{picture}
}
\def\blocgfi{\begin{picture}(80,50)
  \put(60,40){\makebox(0,0){$\rf_i$}}
  \put(60,40){\circle{20}}
  \put(60,30){\vector(0,-1){20}}
  \put(60,0){\circle{20}}
  \put(60,0){\makebox(0,0){$\blue{g}_i$}}
  \put(20,0){\circle{20}}
  \put(20,0){\makebox(0,0){$\epsilon_i$}}
  \put(30,0){\vector(1,0){20}}
\end{picture}
}
\def\blocghfi{\begin{picture}(80,50)
  \put(60,40){\makebox(0,0){$\rf_i$}}
  \put(60,40){\circle{20}}
  \put(60,30){\vector(0,-1){20}}
  \put(65,20){\makebox(0,0){$h$}}
  \put(60,0){\circle{20}}
  \put(60,0){\makebox(0,0){$\blue{g}_i$}}
  \put(20,0){\circle{20}}
  \put(20,0){\makebox(0,0){$\epsilon_i$}}
  \put(30,0){\vector(1,0){20}}
\end{picture}
}
\def\blocgie{\begin{picture}(80,50)
  \put(60,40){\makebox(0,0){$\red{m}$}}
  \put(60,40){\circle{20}}
  \put(60,30){\vector(0,-1){20}}
  \put(60,0){\circle{20}}
  \put(60,0){\makebox(0,0){$\blue{g}_i$}}
  \put(30,0){\circle{20}}
  \put(30,0){\makebox(0,0){$\epsilon_i$}}
  \put(40,0){\vector(1,0){10}}
  \put(0,0){\circle{20}}
  \put(0,0){\makebox(0,0){$\rve$}}
  \put(10,0){\vector(1,0){10}}
\end{picture}
}
\def\blocgievei{\begin{picture}(80,50)
  \put(60,40){\makebox(0,0){$\red{m}$}}
  \put(60,40){\circle{20}}
  \put(60,30){\vector(0,-1){20}}
  \put(60,0){\circle{20}}
  \put(60,0){\makebox(0,0){$\blue{g}_i$}}
  \put(30,0){\circle{20}}
  \put(30,0){\makebox(0,0){$\epsilon_i$}}
  \put(40,0){\vector(1,0){10}}
  \put(0,0){\circle{20}}
  \put(0,0){\makebox(0,0){$\rvei$}}
  \put(10,0){\vector(1,0){10}}
\end{picture}
}
\def\blocfgie{\begin{picture}(80,100)
  \put(60,40){\makebox(0,0){$\rf_i$}}
  \put(60,40){\circle{20}}
  \put(60,30){\vector(0,-1){20}}
  \put(65,20){\makebox(0,0){$\blue{h}$}}
  
  \put(60,0){\circle{20}}
  \put(60,0){\makebox(0,0){$\blue{g}_i$}}
  \put(30,0){\circle{20}}
  \put(30,0){\makebox(0,0){$\epsilon_i$}}
  \put(40,0){\vector(1,0){10}}
  \put(0,0){\circle{20}}
  \put(0,0){\makebox(0,0){$\rve$}}
  \put(10,0){\vector(1,0){10}}
\end{picture}
}
\def\blocvfgie{\begin{picture}(80,100)
  \put(60,80){\makebox(0,0){$\rv_i$}}
  \put(60,80){\circle{20}}
  \put(60,70){\vector(0,-1){20}}
  
  \put(60,40){\makebox(0,0){$\rf_i$}}
  \put(60,40){\circle{20}}
  \put(60,30){\vector(0,-1){20}}
  \put(65,20){\makebox(0,0){$\blue{h}$}}
  
  \put(60,0){\circle{20}}
  \put(60,0){\makebox(0,0){$\blue{g}_i$}}
  \put(30,0){\circle{20}}
  \put(30,0){\makebox(0,0){$\epsilon_i$}}
  \put(40,0){\vector(1,0){10}}
  \put(0,0){\circle{20}}
  \put(0,0){\makebox(0,0){$\rve$}}
  \put(10,0){\vector(1,0){10}}
\end{picture}
}
\def\blocvfhgie{\begin{picture}(80,100)
  \put(60,80){\makebox(0,0){$\rv_i$}}
  \put(60,80){\circle{20}}
  \put(60,70){\vector(0,-1){20}}
  
  \put(60,40){\makebox(0,0){$\rf_i$}}
  \put(60,40){\circle{20}}
  \put(60,30){\vector(0,-1){20}}
  
  \put(30,30){\circle{20}}
  \put(30,30){\makebox(0,0){$\red{h}$}}
  \put(38,22){\vector(1,-1){15}}
  
  \put(60,0){\circle{20}}
  \put(60,0){\makebox(0,0){$\blue{g}_i$}}
  \put(30,0){\circle{20}}
  \put(30,0){\makebox(0,0){$\epsilon_i$}}
  \put(40,0){\vector(1,0){10}}
  \put(0,0){\circle{20}}
  \put(0,0){\makebox(0,0){$\rve$}}
  \put(10,0){\vector(1,0){10}}
\end{picture}
}
\def\blocvfhgiei{\begin{picture}(80,100)
  \put(60,80){\makebox(0,0){$\rv_i$}}
  \put(60,80){\circle{20}}
  \put(60,70){\vector(0,-1){20}}
  
  \put(60,40){\makebox(0,0){$\rf_i$}}
  \put(60,40){\circle{20}}
  \put(60,30){\vector(0,-1){20}}
  
  \put(30,30){\circle{20}}
  \put(30,30){\makebox(0,0){$\red{h}$}}
  \put(38,22){\vector(1,-1){15}}
  
  \put(60,0){\circle{20}}
  \put(60,0){\makebox(0,0){$\blue{g}_i$}}
  \put(30,0){\circle{20}}
  \put(30,0){\makebox(0,0){$\epsilon_i$}}
  \put(40,0){\vector(1,0){10}}
  \put(0,0){\circle{20}}
  \put(0,0){\makebox(0,0){$\rvei$}}
  \put(10,0){\vector(1,0){10}}
\end{picture}
}
\def\blocfgHe{\begin{picture}(80,100)
  \put(60,40){\makebox(0,0){$\rfb$}}
  \put(60,40){\circle{20}}
  \put(60,30){\vector(0,-1){20}}
  \put(65,20){\makebox(0,0){$\blue{\Hb}$}}
  
  \put(60,0){\circle{20}}
  \put(60,0){\makebox(0,0){$\bgb$}}
  \put(30,0){\circle{20}}
  \put(30,0){\makebox(0,0){$\epsilonb$}}
  \put(40,0){\vector(1,0){10}}
  \put(0,0){\circle{20}}
  \put(0,0){\makebox(0,0){$\rve$}}
  \put(10,0){\vector(1,0){10}}
\end{picture}
}
\def\blocvfgHe{\begin{picture}(80,100)
  \put(60,80){\makebox(0,0){$\rvb$}}
  \put(60,80){\circle{20}}
  \put(60,70){\vector(0,-1){20}}
  
  \put(60,40){\makebox(0,0){$\rfb$}}
  \put(60,40){\circle{20}}
  \put(60,30){\vector(0,-1){20}}
  \put(65,20){\makebox(0,0){$\bHb$}}
  
  \put(60,0){\circle{20}}
  \put(60,0){\makebox(0,0){$\bgb$}}
  \put(30,0){\circle{20}}
  \put(30,0){\makebox(0,0){$\epsilonb$}}
  \put(40,0){\vector(1,0){10}}
  \put(0,0){\circle{20}}
  \put(0,0){\makebox(0,0){$\rve$}}
  \put(10,0){\vector(1,0){10}}
\end{picture}
}
\def\blocvfgHei{\begin{picture}(80,100)
  \put(60,80){\makebox(0,0){$\rvb$}}
  \put(60,80){\circle{20}}
  \put(60,70){\vector(0,-1){20}}
  
  \put(60,40){\makebox(0,0){$\rfb$}}
  \put(60,40){\circle{20}}
  \put(60,30){\vector(0,-1){20}}
  \put(65,20){\makebox(0,0){$\bHb$}}
  
  \put(60,0){\circle{20}}
  \put(60,0){\makebox(0,0){$\bgb$}}
  \put(30,0){\circle{20}}
  \put(30,0){\makebox(0,0){$\epsilonb$}}
  \put(40,0){\vector(1,0){10}}
  \put(0,0){\circle{20}}
  \put(0,0){\makebox(0,0){$\rvei$}}
  \put(10,0){\vector(1,0){10}}
\end{picture}
}
\def\blocvfHge{\begin{picture}(80,100)
  \put(60,80){\makebox(0,0){$\rvb$}}
  \put(60,80){\circle{20}}
  \put(60,70){\vector(0,-1){20}}
  
  \put(60,40){\makebox(0,0){$\rfb$}}
  \put(60,40){\circle{20}}
  \put(60,30){\vector(0,-1){20}}
  
  \put(30,30){\circle{20}}
  \put(30,30){\makebox(0,0){$\rHb$}}
  \put(38,22){\vector(1,-1){15}}
  
  \put(60,0){\circle{20}}
  \put(60,0){\makebox(0,0){$\bgb$}}
  \put(30,0){\circle{20}}
  \put(30,0){\makebox(0,0){$\epsilonb$}}
  \put(40,0){\vector(1,0){10}}
  \put(0,0){\circle{20}}
  \put(0,0){\makebox(0,0){$\rve$}}
  \put(10,0){\vector(1,0){10}}
\end{picture}
}
\def\blocvfHgei{\begin{picture}(80,100)
  \put(60,80){\makebox(0,0){$\rvb$}}
  \put(60,80){\circle{20}}
  \put(60,70){\vector(0,-1){20}}
  
  \put(60,40){\makebox(0,0){$\rfb$}}
  \put(60,40){\circle{20}}
  \put(60,30){\vector(0,-1){20}}
  
  \put(30,30){\circle{20}}
  \put(30,30){\makebox(0,0){$\rHb$}}
  \put(38,22){\vector(1,-1){15}}
  
  \put(60,0){\circle{20}}
  \put(60,0){\makebox(0,0){$\bgb$}}
  \put(30,0){\circle{20}}
  \put(30,0){\makebox(0,0){$\epsilonb$}}
  \put(40,0){\vector(1,0){10}}
  \put(0,0){\circle{20}}
  \put(0,0){\makebox(0,0){$\rvei$}}
  \put(10,0){\vector(1,0){10}}
\end{picture}
}
\def\blocvhFge{\begin{picture}(80,100)
  \put(60,80){\makebox(0,0){$\rvb$}}
  \put(60,80){\circle{20}}
  \put(60,70){\vector(0,-1){20}}
  
  \put(60,40){\makebox(0,0){$\rhb$}}
  \put(60,40){\circle{20}}
  \put(60,30){\vector(0,-1){20}}
  
  \put(30,30){\circle{20}}
  \put(30,30){\makebox(0,0){$\bFb$}}
  \put(38,22){\vector(1,-1){15}}

  \put(60,0){\circle{20}}
  \put(60,0){\makebox(0,0){$\bgb$}}
  \put(30,0){\circle{20}}
  \put(30,0){\makebox(0,0){$\epsilonb$}}
  \put(40,0){\vector(1,0){10}}
  \put(0,0){\circle{20}}
  \put(0,0){\makebox(0,0){$\rve$}}
  \put(10,0){\vector(1,0){10}}
\end{picture}
}
\def\blocvhFgei{\begin{picture}(80,100)
  \put(60,80){\makebox(0,0){$\rvb$}}
  \put(60,80){\circle{20}}
  \put(60,70){\vector(0,-1){20}}
  
  \put(60,40){\makebox(0,0){$\rhb$}}
  \put(60,40){\circle{20}}
  \put(60,30){\vector(0,-1){20}}
  
  \put(30,30){\circle{20}}
  \put(30,30){\makebox(0,0){$\bFb$}}
  \put(38,22){\vector(1,-1){15}}

  \put(60,0){\circle{20}}
  \put(60,0){\makebox(0,0){$\bgb$}}
  \put(30,0){\circle{20}}
  \put(30,0){\makebox(0,0){$\epsilonb$}}
  \put(40,0){\vector(1,0){10}}
  \put(0,0){\circle{20}}
  \put(0,0){\makebox(0,0){$\rvei$}}
  \put(10,0){\vector(1,0){10}}
\end{picture}
}
\def\blochFge{\begin{picture}(80,40)
  \put(60,40){\makebox(0,0){$\rhb$}}
  \put(60,40){\circle{20}}
  \put(60,30){\vector(0,-1){20}}
  
  \put(30,30){\circle{20}}
  \put(30,30){\makebox(0,0){$\bFb$}}
  \put(38,22){\vector(1,-1){15}}

  \put(60,0){\circle{20}}
  \put(60,0){\makebox(0,0){$\bgb$}}
  \put(30,0){\circle{20}}
  \put(30,0){\makebox(0,0){$\epsilonb$}}
  \put(40,0){\vector(1,0){10}}
  \put(0,0){\circle{20}}
  \put(0,0){\makebox(0,0){$\rve$}}
  \put(10,0){\vector(1,0){10}}
\end{picture}
}
\def\blochFgei{\begin{picture}(80,40)
  \put(60,40){\makebox(0,0){$\rhb$}}
  \put(60,40){\circle{20}}
  \put(60,30){\vector(0,-1){20}}
  
  \put(30,30){\circle{20}}
  \put(30,30){\makebox(0,0){$\bFb$}}
  \put(38,22){\vector(1,-1){15}}

  \put(60,0){\circle{20}}
  \put(60,0){\makebox(0,0){$\bgb$}}
  \put(30,0){\circle{20}}
  \put(30,0){\makebox(0,0){$\epsilonb$}}
  \put(40,0){\vector(1,0){10}}
  \put(0,0){\circle{20}}
  \put(0,0){\makebox(0,0){$\rvei$}}
  \put(10,0){\vector(1,0){10}}
\end{picture}
}
\def\blocfHge{\begin{picture}(80,40)  
  \put(60,40){\makebox(0,0){$\bfb$}}
  \put(60,40){\circle{20}}
  \put(60,30){\vector(0,-1){20}}
  
  \put(30,30){\circle{20}}
  \put(30,30){\makebox(0,0){$\rHb$}}
  \put(38,22){\vector(1,-1){15}}
  
  \put(60,0){\circle{20}}
  \put(60,0){\makebox(0,0){$\bgb$}}
  \put(30,0){\circle{20}}
  \put(30,0){\makebox(0,0){$\epsilonb$}}
  \put(40,0){\vector(1,0){10}}
  \put(0,0){\circle{20}}
  \put(0,0){\makebox(0,0){$\rve$}}
  \put(10,0){\vector(1,0){10}}
\end{picture}
}
\def\blocfHgei{\begin{picture}(80,40)  
  \put(60,40){\makebox(0,0){$\bfb$}}
  \put(60,40){\circle{20}}
  \put(60,30){\vector(0,-1){20}}
  
  \put(30,30){\circle{20}}
  \put(30,30){\makebox(0,0){$\rHb$}}
  \put(38,22){\vector(1,-1){15}}
  
  \put(60,0){\circle{20}}
  \put(60,0){\makebox(0,0){$\bgb$}}
  \put(30,0){\circle{20}}
  \put(30,0){\makebox(0,0){$\epsilonb$}}
  \put(40,0){\vector(1,0){10}}
  \put(0,0){\circle{20}}
  \put(0,0){\makebox(0,0){$\rvei$}}
  \put(10,0){\vector(1,0){10}}
\end{picture}
}
\def\blocFgHe{\begin{picture}(80,40)
  \put(60,40){\makebox(0,0){$\bfb$}}
  \put(60,40){\circle{20}}
  \put(60,30){\vector(0,-1){20}}
  \put(65,20){\makebox(0,0){$\blue{\rHb}$}}
  
  \put(60,0){\circle{20}}
  \put(60,0){\makebox(0,0){$\bgb$}}
  \put(30,0){\circle{20}}
  \put(30,0){\makebox(0,0){$\epsilonb$}}
  \put(40,0){\vector(1,0){10}}
  \put(0,0){\circle{20}}
  \put(0,0){\makebox(0,0){$\rve$}}
  \put(10,0){\vector(1,0){10}}
\end{picture}
}

\def\ftrfi{\bg(r,\phi)}
\def\wth{\widehat{\tilde{h}}}
\def\varx#1#2{\mbox{Var}_{#1}\left\{#2\right\}}
\def\rvbxi{\rvb_{\xi}}
\def\rVbxi{\rVb_{\xi}}
\def\rvbxih{\rvbh_{\xi}}

\def\repsilon{\red{\epsilon}}

\def\vzj{v_{z_j}}
\def\vxij{v_{\xi_j}}
\def\vei{v_{\epsilon_i}}

\def\vzjh{\widehat{v}_{z_j}}
\def\vxijh{\widehat{v}_{\xi_j}}
\def\veih{\widehat{v}_{\epsilon_i}}

\def\vbeh{\widehat{\vb}_{\epsilon}}
\def\vbxih{\widehat{\vb}_{\xi}}
\def\vbzh{\widehat{\vb}_{z}}

\def\Vbeh{\widehat{\Vb}_{\epsilon}}
\def\Vbxih{\widehat{\Vb}_{\xi}}
\def\Vbzh{\widehat{\Vb}_{z}}

\def\Ybeh{\widehat{\Yb}_{\epsilon}}
\def\Ybxih{\widehat{\Yb}_{\xi}}
\def\Ybzh{\widehat{\Yb}_{z}}

\def\vf{v_f}
\def\vbe{\vb_{\epsilon}}
\def\vbz{\vb_z}
\def\vxi{v_{\xi}}
\def\vbxi{\vb_{\xi}}
\def\Vbe{\Vb_{\epsilon}}
\def\Vbxi{\Vb_{\xi}}
\def\Vbz{\Vb_z}

\def\half{\frac{1}{2}}

\def\dpdx#1#2{\frac{\partial {#1}}{\partial {#2}}}
\def\dpdxd#1#2{\frac{\partial^2 {#1}}{\partial {#2}^2}}
\def\dpdxdy#1#2#3{{{\partial ^2 {#1}}\over{\partial {#2} \partial {#3}}}}

\def\KL#1#2{\mbox{KL}\left[#1 : #2\right]}
\def\Covx#1#2{\mbox{Cov}_{#1}(#2)}
\def\braket#1#2{<{#1}(#2)>}

\def\rlambdab{\red{\lambdab}}
\def\bve{\blue{\ve}}
\def\ralphafjt{\red{\tilde{\alpha}_{f_j}}}
\def\rbetafjt{\red{\tilde{\beta}_{f_j}}}
\def\rSigmabt{\red{\tilde{\Sigmab}}}
\def\rfjt{\red{\tilde{f_j}}}
\def\zetab{\bm{\zeta}}
\def\bvxi{\blue{v_\xi}}
\def\bgamma{\blue{\gamma}}
\def\rgamma{\red{\gamma}}
\def\alphagamz{\alpha_{\gamma_0}}
\def\betagamz{\beta_{\gamma_0}}
\def\alphazetaz{\alpha_{\zeta_0}}
\def\betazetaz{\beta_{\zeta_0}}

\def\bvf{\blue{v_f}}
\def\bvzeta{\blue{v_\zeta}}
\def\rgb{\red{\gb}}
\def\rvxii{\red{v_{\xi_i}}}
\def\rvzeta{\red{v_{\zeta}}}

\usepackage{tikz,mathpazo}
\usetikzlibrary{shapes.geometric, arrows}
\usetikzlibrary{positioning,shapes,shadows,arrows}
\usetikzlibrary{mindmap,backgrounds}
\tikzstyle{startstop} = [rectangle, rounded corners, minimum width=1.5cm, minimum height=0.5cm,text centered, draw=black]
\tikzstyle{io} = [trapezium, trapezium left angle=70, trapezium right angle=110, minimum width=1.5cm, minimum height=0.6cm, text centered, draw=black]
\tikzstyle{process} = [rectangle, minimum width=1cm, minimum height=0.5cm, text centered, draw=black]
\tikzstyle{note} = [text centered]
\tikzstyle{decision} = [diamond, minimum width=1.5cm, minimum height=0.3cm, text centered, draw=black]
\tikzstyle{arrow} = [thick,->,>=stealth]

\def\blambda{\blue{\lambda}}
\def\blambdab{\blue{\lambdab}}

\def\fxy{f(x,y)}
\def\afxy{a\left(f(x,y)\right)}

\def\hxy{h(x,y)}
\def\gxy{g(x,y)}

\def\gt{\widetilde{g}}
\def\ft{\widetilde{f}}
\def\htxy{\widetilde{h}(x,y)}
\def\htzero{\widetilde{h}_0}
\def\htone{\widetilde{h}_1}
\def\httwo{\widetilde{h}_2}

\def\hdxy{h_d(x,y)}
\def\gdxy{g_d(x,y)}

\def\exy{\epsilon(x,y)}
\def\nxy{n(x,y)}
\def\rxy{r(x,y)}

\def\dxy{d(x,y)}
\def\dhxy{d_h(x,y)}
\def\dfxy{d_f(x,y)}

\def\fhxy{\hat{f}(x,y)}
\def\ghxy{\hat{f}(x,y)}
\def\ehxy{\hat{\epsilon}(x,y)}
\def\nhxy{\hat{n}(x,y)}

\def\ftuv{\widetilde{f}(u,v)}
\def\htuv{\widetilde{h}(u,v)}
\def\gtuv{\widetilde{g}(u,v)}
\def\rtuv{\widetilde{r}(u,v)}
\def\ntuv{\widetilde{n}(u,v)}
\def\dtuv{\widetilde{d}(u,v)}
\def\dhtuv{\widetilde{d}_h(u,v)}
\def\dftuv{\widetilde{d}_f(u,v)}

\def\htcuv{\widetilde{h}^*(u,v)}
\def\ftcuv{\widetilde{f}^*(u,v)}

\def\ftuv{\widetilde{f}(u,v)}
\def\htuv{\widetilde{h}(u,v)}
\def\etuv{\widetilde{\epsilon}(u,v)}

\def\ftcuv{\widetilde{f}^*(u,v)}
\def\htcuv{\widetilde{h}^*(u,v)}
\def\etcuv{\widetilde{\epsilon}^*(u,v)}

\def\fhtuv{\widehat{\widetilde{f}}(u,v)}
\def\hhtuv{\widehat{\widetilde{h}}(u,v)}

\def\fhtxy{\widehat{\widetilde{f}}(x,y)}
\def\hhtxy{\widehat{\widetilde{h}}(x,y)}

\def\fhtcuv{\widehat{\widetilde{f}}^*(u,v)}
\def\hhtcuv{\widehat{\widetilde{h}}^*(u,v)}
\def\ghtcuv{\widehat{\widetilde{g}}^*(u,v)}

\def\htzerouv{\widetilde{h}_0(u,v)}
\def\htoneuv{\widetilde{h}_1(u,v)}
\def\httwouv{\widetilde{h}_2(u,v)}
\def\htzerocuv{\widetilde{h}_0^*(u,v)}
\def\htonecuv{\widetilde{h}_1^*(u,v)}
\def\httwocuv{\widetilde{h}_2^*(u,v)}
\def\gtoneuv{\widetilde{g}_1(u,v)}
\def\gttwouv{\widetilde{g}_2(u,v)}

\def\ftuvk{\widetilde{f}^{(k)}(u,v)}
\def\htuvk{\widetilde{h}^{(k)}(u,v)}

\def\fhxyk{\widehat{f}^{(k)}(x,y)}
\def\hhxyk{\widehat{h}^{(k)}(x,y)}
\def\fhxykp{\widehat{f}^{(k+1)}(x,y)}
\def\hhxykp{\widehat{h}^{(k)}(x,y)}

\def\fhk{\widehat{f}^{(k)}}
\def\hhk{\widehat{h}^{(k)}}
\def\fhkp{\widehat{f}^{(k+1)}}
\def\hhkp{\widehat{h}^{(k)}}

\def\fxyk{f^{(k)}(x,y)}
\def\fxykp{f^{(k+1)}(x,y)}

\def\fkxy{f^{(k)}(x,y)}
\def\fkpxy{f^{(k+1)}(x,y)}

\def\ftuvkp{\widetilde{f}^{(k+1)}(u,v)}
\def\htuvkp{\widetilde{h}^{(k+1)}(u,v)}

\def\hktuv{\widetilde{h}_k(u,v)}
\def\gktuv{\widetilde{g}_k(u,v)}

\def\bfxy{\blue{f}(x,y)}
\def\bhxy{\blue{h}(x,y)}
\def\bgxy{\blue{g}(x,y)}

\def\rfxy{\red{f}(x,y)}
\def\rhxy{\red{h}(x,y)}

\def\rftuv{\widetilde{\red{f}}(u,v)}
\def\rhtuv{\widetilde{\red{h}}(u,v)}

\def\bgtuv{\widetilde{\blue{g}}(u,v)}

\def\alphabh{\widehat{\alphab}}

\def\dfdx#1#2{\frac{\partial #1}{\partial #2}}

\def\zxy{z(x,y)}
\def\Txy{T(x,y)}
\def\fxy{f(x,y)}
\def\gzxy{g_0(x,y)}
\def\gxy{g(x,y)}
\def\phixy{\phi(x,y)}
\def\exy{e(x,y)}
\def\epsxy{\epsilon(x,y)}

\def\zhxy{\hat{z}(x,y)}
\def\Thxy{\hat{T}(x,y)}
\def\gzhxy{\hat{g}_0(x,y)}
\def\fhxy{\hat{f}(x,y)}
\def\phihxy{\hat{\phi}(x,y)}
\def\epshxy{\hat{\epsilon}(x,y)}

\def\rfbnn{\rfb_{\tiny NN}}
\def\bvi{\blue{v_{i}}}

\def\Rem#1{{#1}}
\def\red#1{{#1}}
\def\blue#1{{#1}}

\begin{document}

\begin{frontmatter}

\title{Bayesian Physics-Informed Neural Networks for Inverse Problems (BPINN-IP): Application in Infrared Image Processing}

\author[ISCT]{Ali Mohammad-Djafari\corref{cor1}}
\ead{djafari@ieee.org}
\address[ISCT]{former CNRS, founder of ISCT, Bures-sur-Yvette, France \\ 
\textit{Researcher at Shanfeng, Shaoxing and IDT, Ningbo, China} \\ 
Email: djafari@ieee.org, ORCID: 0000-0003-0678-7759 \\ ~}

\author[IDT]{Ning Chu}
\address[IDT]{IDT, Ningbo, China \\ 
Email: chuning1983@sina.com \\ ~}

\author[CSU]{Li Wang}
\address[CSU]{Central South University (CSU), 
Changsha, China \\
Email: wang.li@csu.edu.cn}

\cortext[cor1]{Corresponding author.}

\begin{abstract}
Inverse problems arise across scientific and engineering domains, where the goal is to infer hidden parameters or physical fields from indirect and noisy observations. Classical approaches, such as variational regularization and Bayesian inference, provide well–established theoretical foundations for handling ill-posedness. However, these methods often become computationally restrictive in high-dimensional settings or when the forward model is governed by complex physics.

Physics-Informed Neural Networks (PINNs) have recently emerged as a promising framework for solving inverse problems by embedding physical laws directly into the training process of neural networks. In this paper, we introduce a new perspective on the Bayesian Physics-Informed Neural Network (BPINN) framework, extending classical PINNs by explicitly incorporating training data generation, modeling and measurement uncertainties through Bayesian prior modeling and doing inference with the posterior laws. Also, as we focus on the inverse problems, we call this method BPINN-IP, and we show that the standard PINN formulation naturally appears as its special case corresponding to the Maximum A Posteriori (MAP) estimate.

This unified formulation allows simultaneous exploitation of physical constraints, prior knowledge, and data-driven inference, while enabling uncertainty quantification through posterior distributions. To demonstrate the effectiveness of the proposed framework, we consider inverse problems arising in infrared image processing, including deconvolution and super-resolution, and present results on both simulated and real industrial data.
\end{abstract}


\end{frontmatter}


\section{Introduction}

Inverse problems are fundamental in many areas of science and engineering, where the objective is to infer hidden physical quantities from indirect, noisy, or incomplete observations \cite{Arridge1999, Bodin2009, Luo2014, Miya2002}. Applications include medical and biological imaging \cite{Jin2017, Ker2018}, geophysical exploration \cite{Waheed2021, Puzyrev2019}, and industrial nondestructive evaluation \cite{Miya2002}. 

Traditional approaches to inverse problems rely on analytical formulations, variational regularization, and Bayesian inference \cite{engl1996regularization, Tanner2019, AMD2021}. Bayesian inference is particularly appealing because it explicitly accounts for uncertainties by modeling the posterior distribution of the unknowns conditioned on observations \cite{Jiang2017, AMD2021}. However, classical Bayesian methods such as Markov Chain Monte Carlo (MCMC) or Hamiltonian Monte Carlo can become computationally prohibitive in high-dimensional contexts or when the forward operator is governed by complex physics.

Deep learning (DL) and neural networks (NNs) have recently opened new directions for solving inverse problems \cite{Raissi2019, Yang2021}. Data-driven architectures can approximate linear or nonlinear mappings between observations and unknowns, but they typically require large labeled datasets and often lack interpretability. Physics-Informed Neural Networks (PINNs) address these limitations by embedding the governing physical laws as soft constraints in the loss function, enabling improved generalization and reduced training data requirements \cite{Raissi2019, Haghighat2021}.

This paper advances the PINN paradigm by introducing a new Bayesian formulation specifically for inverse problems, referred to as a Bayesian Physics-Informed Neural Network (BPINN-IP). Although the term BPINN has appeared previously \cite{Yang2021}, our formulation differs in explicitely formulating the uncertainties in the training data generation or their acquisition, as well as prior modeling, which result to obtaining a posterior law for the output of the NN, from which, we can infer on it. In the proposed methodology, we begin by describing a probabilistic generative model that characterizes both the training data and the observation model. We then impose the physical forward operator as a constraint within the Bayesian inference framework, resulting in posterior distributions over the unknown physical fields as well as the neural network parameters.

This formulation unifies deterministic PINNs and Bayesian inference: the classical PINN loss arises as the Maximum A Posteriori (MAP) estimator of the BPINN-IP, while the full BPINN-IP enables both estimates and uncertainty quantification (UQ). The remainder of this paper details the proposed probalilistic formulation, demonstrates its performance through synthetic and real-world infrared imaging examples, and discusses implications for broader classes of inverse problems.

\section{Literature Review and State of the Art in NN, PINN, and BPINN Methods for Inverse Problems}

Inverse problems have been studied extensively using analytical, variational, and probabilistic approaches. Analytical methods provide exact solutions but are limited to simple, well-posed configurations \cite{engl1996regularization}. Variational regularization techniques—such as Tikhonov regularization \cite{Tanner2019} and sparsity-promoting priors—are widely used to stabilize ill-posed character of inverse problems \cite{Slagel2019}, though selecting appropriate regularization parameters remains challenging, especially in noisy or heterogeneous environments.

Bayesian inference offers a principled framework for incorporating prior knowledge and quantifying uncertainties \cite{AMD2021}. Sampling-based strategies, such as Markov Chain Monte Carlo (MCMC) \cite{Beskos2015} and Hamiltonian Monte Carlo \cite{Fichtner2019}, allow exploration of complex posterior distributions but often suffer from high computational cost. Variational approximate inference methods provide more scalable approximations, yet may struggle to capture multimodal or high-dimensional posterior geometries.

Deep learning has significantly advanced inverse problem-solving capabilities via its capability of doing fast inference. Convolutional (CNN) and recurrent neural networks (RNN) learn complex nonlinear mappings from data \cite{Jin2017, Puzyrev2019}, but their dependence on large labeled datasets and limited interpretability can hinder physical consistency. Physics-Informed Neural Networks (PINNs) address these challenges by incorporating governing equations as soft constraints in the optimization problem, improving generalization and reducing data requirements \cite{Raissi2019}.

Recent developments have extended PINNs to Bayesian formulations (often termed BPINNs), introducing stochastic regularization, probabilistic priors, or approximate posterior inference \cite{Yang2021}. While these approaches are promising, significant gaps remain in terms of training robustness, integration of forward-model uncertainties, and validation on real-world experimental datasets.

In this work, we build on these foundations by proposing a new Bayesian PINN framework specifically tailored for inverse problems, that we call (BPINN-IP). The method begins with a generative prior model describing the training data, proceeds with a Bayesian interpretation of the training process, and incorporates the forward physical model to derive posterior laws for both the unknown physical quantities and the neural network parameters. We show explicitly that deterministic PINNs arise as a special case corresponding to the MAP solution of the BPINN-IP. The resulting framework combines the interpretability and physical consistency of PINNs with the principled uncertainty quantification of Bayesian inference, yielding a robust and general methodology applicable to both simulated and experimental scenarios.

\section{Proposed Bayesian PINN-IP framework}
We start by introducing the proposed Bayesian Physics-Informed Neural Network for inverse problems (BPINN-IP) framework in a sequence of five conceptual steps.  

\ben
\item 
\textbf{Forward (generative) model.}\\ 
We start by assuming a general forward or generative model
\beq
\bgb = \Hb(\rfb) + \epsilonb,
\label{eq01}
\eeq
where $\rfb$ is the input (unknown), $\bgb$ the output (observed data), $\Hb$ the forward model, and $\epsilonb$ the (additive) modeling and measurement errors. In the linear case, this simplifies to
\beq
\bgb = \Hb \rfb + \epsilonb,
\label{eq02}
\eeq
where $\Hb$ is now a matrix. In many imaging inverse problems, $\rfb \in \mathbb{R}^N$ is a vector containing all pixels of the unknown image, and $\bgb \in \mathbb{R}^M$ collects all measured data (which may itself be an image). Thus, $\Hb \in \mathbb{R}^{M \times N}$. 

In realistic imaging problems, $M$ and $N$ can be very large (e.g.\ for images of size $m \times n$, we have $M=N=m\times n$), making it infeasible to store the whole matrix $\Hb$ explicitly. Fortunately, practical algorithms only need the forward operator $\rfb \mapsto \Hb\rfb$ and its adjoint $\bgb \mapsto \Hb^T\bgb$, which can often be implemented efficiently without forming explicitely $\Hb$ explicitly.

\item 
\textbf{Prior modeling and Bayesian formulation.}\\ 
We now assign priors $p(\rfb)$ and $p(\epsilonb)$. From the latter and the forward model, we obtain the likelihood $p(\bgb|\rfb)$. Bayes’ rule then gives the posterior
\beq
p(\rfb|\bgb) = \frac{p(\bgb|\rfb)\, p(\rfb)}{p(\bgb)},
\label{eq03}
\eeq
where $p(\bgb|\rfb) = p(\epsilonb)$ with $\epsilonb = \bgb - \Hb\rfb$, and
\[
p(\bgb) = \int p(\bgb|\rfb)\, p(\rfb)\,\mathrm{d}\rfb.
\]
For the class of linear inverse problems
$
\bgb = \Hb\rfb + \epsilonb
$
with Gaussian likelihood
\[
p(\bgb|\rfb) = \Nc(\bgb \mid \Hb\rfb,\bve\Ib),
\]
and Gaussian prior
\[
p(\rfb) = \Nc(\rfb \mid \bar{\fb},\bvf\Ib),
\]
it is well known (and easy to show) that the posterior $p(\rfb|\bgb)$ is also Gaussian:
\beq
p(\rfb|\bgb) = \Nc(\rfb \mid \rfbh,\rSigmabh)
\label{eq04}
\eeq
with
\beq
\left\{\barr{l}
\rfbh = [\Hb^T\Hb + \lambda\Ib]^{-1}\big(\Hb^T\bgb + \lambda\bar{\fb}\big),\\[6pt]
\rSigmabh = \bve\, [\Hb^T\Hb + \lambda\Ib]^{-1}, \qquad \lambda = \dfrac{\bve}{\bvf}.
\earr\right.
\label{eq05} 
\eeq
The mean $\rfbh$ corresponds to the classical Tikhonov solution, and $\rSigmabh$ quantifies its uncertainty.

\item 
\textbf{Neural network surrogate for the inverse map.} \\ 
Next, we design an appropriate neural network that takes $\bgb$ as input and outputs an estimate $\rfb_{NN}$ of $\rfb$. This NN can be viewed as an approximate inverse or surrogate:
\[
\bgb \;\Ra\; \fbox{~NN ($\rwb$)~} \;\Ra\; \rfb_{NN}~~,
\]
where $\rwb$ denotes the NN parameters. The crucial task is to choose an appropriate network architecture: input layer, number and sizes of hidden layers, output layer, connections, activation functions, and thus the dimension of the parameter vector $\rwb$. In many imaging applications, U-Net-like architectures are particularly convenient, but the proposed framework is agnostic to the specific choice.

\item 
\textbf{Training the NN: supervised and unsupervised cases.} \\ 
We distinguish two main training scenarios.

\bit
\item \textbf{Supervised case.} Here, we have a set of training pairs $\{\bgb_{Ti}, \bfb_{Ti}\}$, where $\bfb_{Ti}$ are reference (ground-truth) images. The training step estimates the NN parameters $\rwbh$, schematically:
\[
\btabu{@{}c@{}} \text{Data} \\ $\{\bgb_{Ti}, \bfb_{Ti}\}$ \etabu
\;\Ra\;
\fbox{\btabu{c} Training step \\ NN ($\rwb$) \etabu}
\;\Ra\; \rwbh.
\]
\item \textbf{Unsupervised case.} Here, only the degraded data $\{\bgb_{Ti}\}$ are available. Classical NNs cannot handle this situation directly, whereas the PINN approach, thanks to the physics-informed part of the loss, can still be trained and yield an estimate $\rwbh$:
\[
\btabu{@{}c@{}} \text{Data} \\ $\{\bgb_{Ti}\}$\etabu
\;\Ra\;
\fbox{\btabu{c} Training step \\ NN ($\rwb$) \etabu}
\;\Ra\; \rwbh.
\]
\eit

The index $i$ enumerates the samples in the training dataset. At the end of training, we obtain an estimate $\rwbh$ of the NN parameters. The training step is generally the most computationally demanding phase and requires careful choice of the optimization criterion and algorithm.

In standard supervised regression, the loss is the classical output-error loss
\[
J_{NN}(\rwb) = \sum_i \|\rfb_{NNi}(\rwb) - \rfb_{Ti}\|^2,
\]
where $\rfb_{NNi}(\rwb)$ is the NN output for input $\bgb_{Ti}$, and $\rfb_{Ti}$ is the corresponding ground truth. Later, additional terms will be added to this loss, either for regularization or to enforce physics-informed constraints.

After defining the loss, we choose an optimization algorithm, typically gradient-based (stochastic gradient descent and its variants), along with hyperparameters such as learning rate, number of iterations, and stopping criteria. Hyperparameter tuning is often necessary to obtain good generalization and to avoid overfitting.

\item 
\textbf{Inference and uncertainty quantification.}
Once the NN has been trained, we use it to infer $\rfb$ for new data $\bgb_j$:
\[
\bgb_j \;\Ra\; \fbox{\btabu{c} Inference step \\ NN ($\rwbh$) \etabu} \;\Ra\; \rfbh_{NNj}~~.
\]
The hope is that $\rfbh_{NNj}$ is close to the (unknown) ground truth $\rfb_j$, for example in terms of
\[
\Delta = \|\rfbh_{NNj} - \rfb_j\|^2.
\]
Beyond a point estimate, we would like to quantify the uncertainty of $\rfbh_{NNj}$. The inference step with uncertainty quantification (UQ) can be summarized as
\[
\bgb_j \;\Ra\; \fbox{\btabu{c} Inference step \\ with UQ \\ NN ($\rwbh$)\etabu}
\;\Ra\;
\left\{\barr{c} \rfbh_{NNj} \\[6pt] \Sigmabh_{NNj} \earr\right.,
\]
where $\rfbh_{NNj} = \mathbb{E}[\rfb_j|\bgb_j]$ is the posterior mean, and $\Sigmabh_{NNj}$ its covariance. As $\rfb \in \mathbb{R}^N$ is high dimensional, $\Sigmabh_{NNj} \in \mathbb{R}^{N\times N}$ is even larger, and in practice we often restrict attention to its diagonal (pixelwise variances). For imaging problems, one can thus visualize both a mean image and a variance (or standard deviation) image; examples will be provided in the Results section.
\een

To go into more detail, we now discuss separately the supervised and unsupervised cases.

\subsection{Supervised training data}
We first consider the supervised case, where a labeled training dataset $\{\bgb_{Ti}, \bfb_{Ti}\}$ is available. The objective is to estimate the NN parameters $\rwbh$ using these paired data:
\[
\btabu{@{}c@{}} \text{Data} \\ $\{\bgb_{Ti}, \bfb_{Ti}\}$\etabu
\;\Ra\;
\fbox{\btabu{c} Supervised Training step \\ NN ($\rwb$) \etabu}
\;\Ra\; \rwbh~~.
\]

Inside the supervised training block, the flow of variables can be illustrated as
\[
\btabu{@{}c@{}} \text{Data} \\ $\{\bgb_{Ti}, \bfb_{Ti}\}$\etabu
\;\Ra\;
\fbox{~NN ($\rwb$)~}
\;\Ra\;
\rfb_{NNi}
\;\Ra\;
\fbox{~$\Hb$~}
\;\Ra\;
\rgbh_{NNi}~~,
\]
where $\rfb_{NNi}$ is the NN output corresponding to $\bgb_{Ti}$, and $\rgbh_{NNi} = \Hb \rfb_{NNi}$ is the corresponding prediction in the data space.

The main task is to define a criterion $J(\rwb)$ to be minimized in order to obtain $\rwbh$. In classical NN training, the loss depends only on the discrepancy between NN output $\rfb_{NNi}(\rwb)$ and reference $\bfb_{Ti}$:
\beq
J_{NN}(\rwb) = \sum_i \|\bfb_{Ti} - \rfb_{NNi}(\rwb)\|^2,
\label{eq06}
\eeq
which does not require the forward model $\Hb$.

In PINN methods, the criterion combines the output error and a physics-informed term:
\begin{align}
J_{PI}(\rwb)
&= \sum_i \|\bar{\bfb} - \rfb_{NNi}(\rwb)\|^2 
+ \lambda \|\bgb_{Ti} - \Hb \rfb_{NNi}(\rwb)\|^2,
\label{eq07}
\end{align}
where the first term can represent a data fidelity (or regularization) relative to some reference $\bar{\bfb}$, and the second term enforces consistency with the forward model. One of the main goals of the proposed methodology is to show that such criteria (classical NN and physics-informed PINN) can be recovered as special cases of a more general Bayesian probabilistic formulation, which we detail next.

\bit
\item \textbf{Step 1: probabilistic modeling of the training data.}
We first focus on how the training data are generated. This step explicitly accounts for the forward model $\Hb$ and the uncertainties in the data via a Bayesian approach. We assume that, given the true $\rfb_i$ and the forward model $\Hb$, the observed data $\{\bgb_{Ti}\}$ are conditionally independent:
\beq
p(\{\bgb_{Ti}\}|\{\rfb_i\}) = \prod_i p(\bgb_{Ti}|\rfb_i),
\label{eq08}
\eeq
with, for example, a Gaussian likelihood
\beq
p(\bgb_{Ti}|\rfb_i) = \Nc\big(\bgb_{Ti} \mid \Hb\rfb_i,\bve_i\Ib\big).
\label{eq09}
\eeq
Similarly, for the reference data $\{\bfb_{Ti}\}$, we assume
\beq
p(\{\bfb_{Ti}\}|\{\rfb_i\}) = \prod_i p(\bfb_{Ti}|\rfb_i),
\label{eq10}
\eeq
with
\beq
p(\bfb_{Ti}|\rfb_i) = \Nc\big(\bfb_{Ti} \mid \rfb_i,\bvf_i\Ib\big).
\label{eq11}
\eeq

Here, $\bve_i$ and $\bvf_i$ are the variances governing the data generation process, encoding both the forward model physics and the measurement uncertainties. The process can be summarized schematically as
\[
\rfb_i
\;\Ra\;
\fbox{\btabu{c} Physics-based \\ Generating Data \\ System \etabu}
\barr{l}
\;\Ra\; \bfb_{Ti} \sim p(\bfb_{Ti}|\rfb_i,\bvf_i), \\[4pt]
\;\Ra\; \bgb_{Ti} \sim p(\bgb_{Ti}|\rfb_i,\bve_i),
\earr
\]
where the notation $\xb \sim p(\xb)$ means that $\xb$ is generated from the distribution $p(\xb)$.

We also assign a prior $p(\rfb_i)$ to the true $\rfb_i$, for example a Gaussian prior
\beq
p(\rfb_i|\bar{\bfb},v_i) = \Nc(\rfb_i \mid \bar{\bfb}, v_i\Ib),
\label{eq12}
\eeq
where $\bar{\bfb}$ is a background (mean) image and $v_i$ quantifies the variability of $\rfb_i$ around $\bar{\bfb}$.

\item \textbf{Step 2: posterior of $\rfb_i$ given the training data.}
Using Bayes’ rule, the posterior distribution of $\rfb_i$ given the training data is
\begin{align}
p(\rfb_i|\{\bgb_{Ti}, \bfb_{Ti}\})
&\propto p(\bgb_{Ti}|\rfb_i,\bve_i)\,
p(\bfb_{Ti}|\rfb_i,\bvf_i)\,
p(\rfb_i|\bar{\bfb},v_i).
\label{eq13}
\end{align}
With the Gaussian models in (\ref{eq09}), (\ref{eq11}), and (\ref{eq12}), we can write
\beq
p(\rfb_i|\{\bgb_{Ti},\bfb_{Ti}\}) \propto \expf{-J(\rfb_i)},
\label{eq14}
\eeq
where
\begin{align}
J(\rfb_i) &=
\frac{1}{2\bve_i}\|\bgb_{Ti} - \Hb\rfb_i\|^2 
+ \frac{1}{2\bvf_i} \|\bfb_{Ti} - \rfb_i\|^2
+ \frac{1}{2v_i} \|\rfb_i - \bar{\bfb}\|^2.
\label{eq15}
\end{align}
It follows that the posterior is also Gaussian:
\beq
p(\rfb_i|\{\bgb_{Ti},\bfb_{Ti}\}) = \Nc(\rfb_i \mid \rfbh_i,\rSigmabh_i),
\label{eq16}
\eeq
with the usual linear-Gaussian expressions (here written schematically as)
\beq
\left\{\barr{l}
\disp{
\rfbh_i =
[\Hb^T\Hb + \lambda_i\Ib + \mu_i\Ib]^{-1}
\big(\Hb^T\bgb_{Ti} + \lambda_i\bfb_{Ti} + \mu_i\bar{\rfb}\big)},\\[6pt]
\disp{
\rSigmabh_i =
[\Hb^T\Hb + \lambda_i\Ib + \mu_i\Ib]^{-1}\bve_i,
\quad
\lambda_i = \dfrac{\bve_i}{\bvf_i}, \quad \mu_i = \dfrac{\bve_i}{v_i}.}
\earr\right.
\label{eq17}
\eeq
These relations form the basic building blocks for the proposed BPINN formulation, both at training and inference stages.

\item \textbf{Step 3: Bayesian modeling of the NN training.}
We now consider the training step of the NN, denoting by $\rfb_{NNi}(\rwb)$ the NN output for input $\bgb_{Ti}$ and parameters $\rwb$. The full flow of variables, from physics-based data generation through the NN and back to the data space, is
\[
\fbox{\btabu{@{}c@{}} Generating \\ Data System \etabu}
\;\Ra\;
\left\{\barr{@{}c@{}} \bgb_{Ti}\\ \bfb_{Ti}\earr\right\}
\;\Ra\;
\fbox{NN($\rwb$)}
\;\Ra\;
\rfb_{NNi}
\;\Ra\;
\fbox{$\Hb$}
\;\Ra\;
\rgbh_{NNi}.
\]

Focusing on the training step, where the goal is to estimate $\rwb$, we introduce the following probabilistic modeling:
\beq
\left\{\barr{l}
p(\bfb_{Ti}|\rwb) = \Nc\big(\bfb_{Ti} \mid \rfb_{NNi}(\rwb),\bvf_i\Ib\big),\\[6pt]
p(\bgb_{Ti}|\rwb) = \Nc\big(\bgb_{Ti} \mid \Hb\rfb_{NNi}(\rwb),\bve_i\Ib\big).
\earr\right.
\label{eq18}
\eeq
Using Bayes’ rule with a prior $p(\rwb)$, we obtain the posterior
\beq
p(\rwb | \{\bgb_{Ti},\bfb_{Ti}\}) 
\propto 
\prod_i p(\bfb_{Ti}|\rwb)\, p(\bgb_{Ti}|\rwb)\, p(\rwb),
\label{eq19}
\eeq
which can be rewritten as
\beq
p(\rwb | \{\bgb_{Ti},\bfb_{Ti}\}) \propto \expf{-J(\rwb)},
\label{eq20}
\eeq
with
\begin{align}
J(\rwb)
&=
\sum_i \frac{1}{2\bvf_i} \|\bfb_{Ti} - \rfb_{NNi}(\rwb)\|^2
+ \frac{1}{2\bve_i}\|\bgb_{Ti} - \Hb\rfb_{NNi}(\rwb)\|^2 
- \ln p(\rwb).
\label{eq21}
\end{align}
We can thus identify three components:
\beq
J_{NN}(\rwb) = \sum_i \frac{1}{2\bvf_i} \|\bfb_{Ti} - \rfb_{NNi}(\rwb)\|^2,
\label{eq22}
\eeq 
\begin{align}
J_{PI}(\rwb) = \sum_i \frac{1}{2\bve_i} \|\bgb_{Ti} - \Hb\rfb_{NNi}(\rwb)\|^2,
\label{eq23} 
\end{align}
and the prior-induced term
\beq
J_{PR}(\rwb) = - \ln p(\rwb).
\label{eq24}
\eeq
In other words, the classical NN loss $J_{NN}$, the physics-informed loss $J_{PI}$, and the regularization via the prior $J_{PR}$ all arise from a single Bayesian posterior formulation.

The posterior $p(\rwb | \{\bgb_{Ti},\bfb_{Ti}\})$ in (\ref{eq20})–(\ref{eq21}) can be used to derive either point estimates of $\rwb$ (e.g.\ MAP) or to quantify uncertainty in $\rwb$.

A useful choice for the prior $p(\rwb)$ is a sparsity-enforcing prior \cite{MOHAMMADDJAFARI2015128}, such as
\beq
p(\rwb) \propto \expf{-\gamma_w \|\rwb\|_\beta^\beta},
\label{eq25}
\eeq
with $\beta \in (0,2]$. The special case $\beta = 1$ (Laplace prior) is closely related to \emph{dropout} techniques, which can be interpreted as a Monte Carlo approximation to Bayesian inference in NNs. In practice, computing the full posterior analytically is intractable, especially with nonlinear activation functions, and one typically resorts to approximate inference (e.g.\ variational methods or Monte Carlo dropout).
\eit

\subsection{Unsupervised training data}
In the unsupervised case, only the degraded training data $\{\bgb_{Ti}\}$ are available. Classical supervised NN methods cannot be applied, as there is no ground truth. This is precisely where the physics-informed PINN/BPINN framework is particularly valuable. The schematic data flow is
\[
\btabu{@{}c@{}} \text{Data} \\ $\{\bgb_{Ti}\}$\etabu
\;\Ra\;
\fbox{~NN ($\rwb$)~}
\;\Ra\;
\rfb_{NNi}
\;\Ra\;
\fbox{~$\Hb$~}
\;\Ra\;
\rgbh_{NNi}.
\]
We can still follow the same Bayesian steps as in the supervised case, with appropriate modifications.

\bit
\item \textbf{Data likelihood.}
We assume that, given the true $\rfb_i$ and the forward model $\Hb$, the data $\{\bgb_{Ti}\}$ are conditionally independent and Gaussian:
\begin{align}
p(\{\bgb_{Ti}\}|\{\rfb_i\}) 
&=  \prod_i \Nc\big(\bgb_{Ti}\mid \Hb\rfb_i,\bve_i\Ib\big) \nonumber \\
&\propto \expf{-\sum_i \frac{1}{2\bve_i}\|\bgb_{Ti}-\Hb\rfb_i\|^2}.
\label{eq26}
\end{align}

\item \textbf{Prior on $\rfb_i$.}
We assign a prior $p(\rfb_i)$ to encode prior knowledge (e.g.\ smoothness or sparsity) about the unknowns. For imaging problems, a common choice is a Tikhonov, total variation, or generalized sparsity prior, for example
\beq
p(\rfb_i) \propto \expf{-\gamma\|\Db \rfb_i\|_\beta^\beta},
\label{eq27}
\eeq
where $\Db$ is a finite-difference (or other) operator, and $\beta \in (0,2]$.

\item \textbf{Posterior of $\rfb_i$.}
Using Bayes’ rule, we obtain
\begin{align}
p(\rfb_i|\{\bgb_{Ti}\})
&\propto \expf{-J(\rfb_i)},
\label{eq28}
\end{align}
with
\begin{align}
J(\rfb_i)
&= \sum_i \frac{1}{2\bve_i}\|\bgb_{Ti}-\Hb\rfb_i\|^2
+ \gamma\|\Db\rfb_i\|_\beta^\beta.
\label{eq29}
\end{align}

\item \textbf{Posterior over NN parameters $\rwb$.}
As in the supervised case, we can formulate a posterior over $\rwb$ by passing the data through the NN:
\begin{align}
p(\rwb|\{\bgb_{Ti}\}) 
&\propto \expf{-J(\rwb)}, 
\label{eq30}
\end{align}
with
\begin{align}
J(\rwb)
&= \sum_i \frac{1}{2\bve_i}\|\bgb_{Ti}-\Hb\rfb_{NNi}(\rwb)\|^2 
+ \gamma_w\|\rwb\|_{\beta_w}^{\beta_w},
\label{eq31}
\end{align}
where the first term enforces physics-based consistency in the data space, and the second term imposes a prior (e.g.\ sparsity) on the NN weights. This posterior can then be used to estimate $\rwb$ and quantify uncertainty.
\eit

In both supervised and unsupervised regimes, the practitioner must still choose an appropriate NN architecture (depth, number of hidden units, type of convolutions, etc.) and optimization scheme (learning rate, batch size, stopping criteria, etc.). These choices make the practical implementation of BPINNs nontrivial and often require substantial experimentation. For a discussion on the choice of NN structures in inverse problems, we refer to \cite{psf2023009014,proceedings2019033016}. 

We summarize the two main steps of the proposed method (training) and (Inference and UQ), as they have been implemented, in Algorithm~1 and Algorithm~2. One main remark about the UQ, the whole computation of the posterior covariance is too costly and almost impossible. So, in Algorithm~2, only an approximate computation of the means and variances are implemented through simulated based Bayesian inference and the famous Drop-out technic often used in NNs implementation. 

\clearpage\newpage

\begin{algorithm}[t]
\caption{Bayesian Physics-Informed Neural Network for Inverse Problems (BPINN-IP) training steps.}
\label{alg:BPINN-IP}
\begin{algorithmic}[1]
\Require Forward operator $\Hb$;\\
\hspace*{1.5em} Supervised data $\{\bgb_{Ti}, \bfb_{Ti}\}$ or unsupervised data $\{\bgb_{Ti}\}$;\\
\hspace*{1.5em} Noise variances $\{\bve_i\}$, (optionally $\{\bvf_i\}$ in supervised case);\\
\hspace*{1.5em} Prior hyperparameters $(\gamma_w,\beta_w)$ for $p(\rwb)$; learning rate $\eta$;\\
\hspace*{1.5em} Maximum number of iterations $K$ (or stopping criterion).
\State Initialize NN parameters $\rwb^{(0)} \sim p(\rwb)$
\For{$k = 0,1,\dots,K-1$}
  \State Sample a mini-batch $\mathcal{B} \subset \{1,\dots,N_T\}$ of training indices
  \For{each $i \in \mathcal{B}$}
    \State \textbf{Forward pass in image space:}
           $\rfb_{NNi} \gets \mathrm{NN}(\bgb_{Ti}; \rwb^{(k)})$
    \State \textbf{Forward pass in data space (physics):}
           $\rgbh_{NNi} \gets \Hb \rfb_{NNi}$
    \If{supervised}
      \State Data-fidelity (NN) term:
      \[
      J_{NN}^{(i)} \gets \frac{1}{2\bvf_i}
      \big\|\bfb_{Ti} - \rfb_{NNi}\big\|^2
      \]
    \Else
      \State $J_{NN}^{(i)} \gets 0$ \Comment{no labeled $\bfb_{Ti}$ in unsupervised case}
    \EndIf
    \State Physics-informed term:
    \[
    J_{PI}^{(i)} \gets \frac{1}{2\bve_i}
    \big\|\bgb_{Ti} - \rgbh_{NNi}\big\|^2
    \]
  \EndFor
  \State Prior (regularization) term on NN parameters:
  \[
  J_{PR}(\rwb^{(k)}) \gets - \ln p(\rwb^{(k)})
  \quad
  \text{e.g. }
  J_{PR}(\rwb^{(k)}) = \gamma_w \big\|\rwb^{(k)}\big\|_{\beta_w}^{\beta_w}
  \]
  \State Total mini-batch loss:
  \[
  J(\rwb^{(k)}) \gets
  \sum_{i \in \mathcal{B}}
  \big(J_{NN}^{(i)} + J_{PI}^{(i)}\big) + J_{PR}(\rwb^{(k)})
  \]
  \State Compute gradient
  $\nabla_{\rwb} J(\rwb^{(k)})$ by backpropagation
  \State Update NN parameters (e.g. gradient descent, Adam):
  \[
  \rwb^{(k+1)} \gets
  \rwb^{(k)} - \eta\, \nabla_{\rwb} J(\rwb^{(k)})
  \]
\EndFor
\State \textbf{Output:} trained parameters $\rwbh = \rwb^{(K)}$ approximating the posterior mode (MAP) of $p(\rwb \mid \{\bgb_{Ti},\bfb_{Ti}\})$
\end{algorithmic}
\end{algorithm}

\clearpage\newpage
\begin{algorithm}[t]
\caption{BPINN-IP inference and uncertainty quantification (UQ)}
\label{alg:BPINN_inference}
\begin{algorithmic}[1]
\Require Trained NN parameters $\rwbh$ (from Algorithm~\ref{alg:BPINN-IP});\\
\hspace*{1.5em} Forward operator $\Hb$;\\
\hspace*{1.5em} New observed data $\bgb_j$;\\
\hspace*{1.5em} Number of Monte Carlo samples $T$ (e.g.\ $T=20$--$50$).
\State \textbf{Initialization:}
\Statex \quad Set $t \gets 1$;
          initialize storage $\{\rfb_{NNj}^{(t)}\}_{t=1}^T$ (and optionally $\{\rgbh_{NNj}^{(t)}\}_{t=1}^T$).
\For{$t = 1,\dots,T$}
  \State Activate stochastic components of NN (e.g.\ dropout) using $\rwbh$
  \State \textbf{Forward pass in image space:}
         \[
         \rfb_{NNj}^{(t)} \gets \mathrm{NN}(\bgb_j; \rwbh, \text{stochastic})
         \]
  \State \textbf{Optional forward pass in data space (physics check):}
         \[
         \rgbh_{NNj}^{(t)} \gets \Hb\, \rfb_{NNj}^{(t)}
         \]
\EndFor
\State \textbf{Posterior mean estimate (image space):}
\[
\rfbh_{NNj} \;\gets\; 
\frac{1}{T} \sum_{t=1}^T \rfb_{NNj}^{(t)}
\]
\State \textbf{Posterior variance (pixelwise) estimate:}
\[
\Sigmabh_{NNj} \;\gets\; 
\frac{1}{T-1} \sum_{t=1}^T 
\big(\rfb_{NNj}^{(t)} - \rfbh_{NNj}\big)
\odot
\big(\rfb_{NNj}^{(t)} - \rfbh_{NNj}\big)
\]
where $\odot$ denotes elementwise (Hadamard) product.
\State \textbf{Optional data-space consistency check:}
\[
\gbh_{NNj} \;\gets\;
\frac{1}{T} \sum_{t=1}^T \rgbh_{NNj}^{(t)},
\qquad
\text{and compute } \|\bgb_j - \gbh_{NNj}\|^2
\]
as a physics-based quality indicator.
\State \textbf{Output:} posterior mean image $\rfbh_{NNj}$ and variance (or standard deviation) map $\Sigmabh_{NNj}$, which can be visualized as an \emph{expected value image} and an \emph{uncertainty image}.
\end{algorithmic}
\end{algorithm}

\clearpage\newpage

\newpage
\section{Application in infrared image processing}
We consider two representative applications in infrared (IR) imaging: 
(i) infrared image restoration (deblurring), and 
(ii) infrared image super-resolution. 
Both are classical inverse problems in imaging science and computer vision 
\cite{5413589,Humblot2006}, and both can be written in the general form 
$
\bgb = \Hb \rfb + \epsilonb
$
with appropriate modeling of the imaging physics.

\subsection{Infrared image restoration}

In this application, we consider a simplified but representative IR forward model:
\beq
g(x,y) = h(x,y) * \phi(f(x,y)) + \epsilon(x,y),
\label{eq32}
\eeq
where $g(x,y)$ is the observed IR image, $h(x,y)$ is the point spread function (PSF) of the imaging system (accounting for heat diffusion in air and the optical response of the IR camera), $f(x,y)$ is the unknown surface temperature distribution, and $\epsilon(x,y)$ represents modeling and measurement errors.

The nonlinear pointwise mapping $\phi(\cdot)$ models the combined effects of:
(1) surface emissivity, 
(2) ambient humidity and temperature, 
(3) background radiance, and 
(4) attenuation due to the distance between the surface and the sensor.
More advanced emissivity–radiation models can be incorporated when available (see \cite{mohammaddjafari2024modelbasedphysicsinformed}).

\begin{figure}[h]
\[
\hspace{-5mm}
\btabu{@{}c@{}} $\rfxy$ \\ \includegraphics[scale=.2]{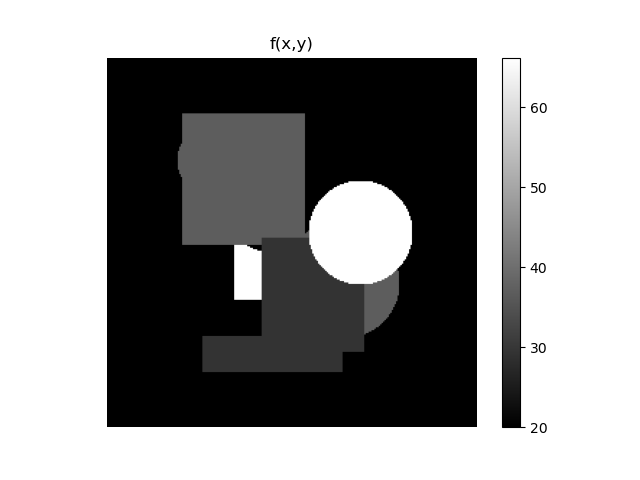} \etabu
\hspace{-5mm}
\small{
\fbox{\btabu{@{}c@{}} Emissivity \\ Environment \\ effects \\ $\phi(f)$ \etabu}
}
\btabu{@{}c@{}} $\phi(x,y)$ \\ \includegraphics[scale=.2]{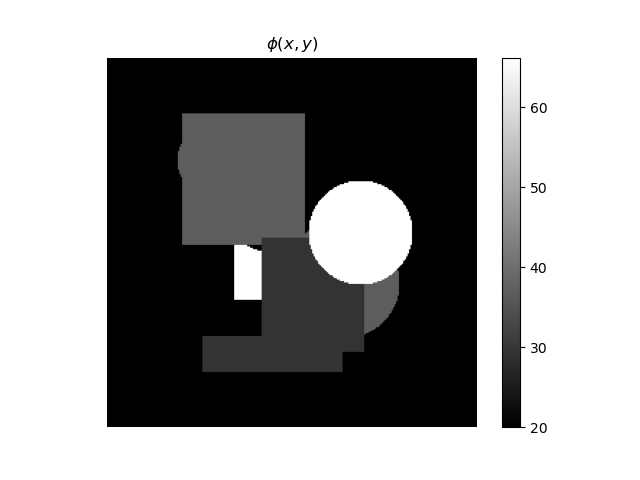} \etabu
\hspace{-5mm}
\small{
\fbox{\btabu{@{}c@{}} Diffusion \\ Convolution \\ $g = h * \phi$ \etabu}
}
\btabu{@{}c@{}} $\bgxy$ \\ \includegraphics[scale=.2]{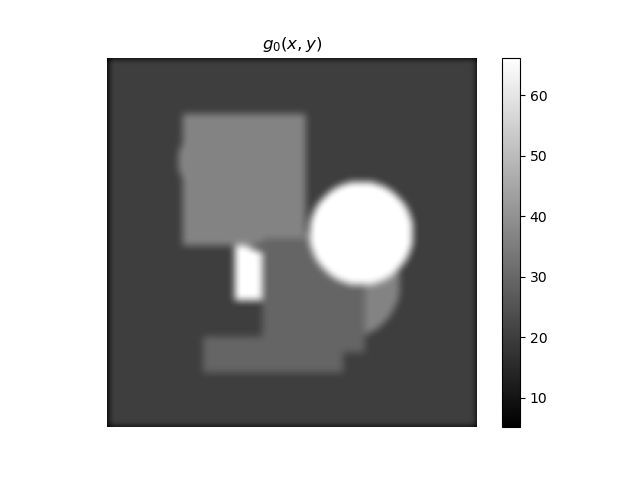} \etabu
\]
\caption{Infrared simplified forward model: from left to right. True temperature distribution $\fxy$; emissivity and environmental mapping $\phi(f)$; diffusion–convolution via the PSF $h$; observed IR image $g$.}
\label{fig01}
\end{figure}

Figure~\ref{fig01} illustrates these steps: nonlinear emissivity correction, diffusion–convolution, and final IR measurement.

After discretization, the model becomes:
\beq
\bgb = \Hb \Phib \rfb + \epsilonb,
\label{eq33}
\eeq
where $\bgb$ contains all IR image pixels, $\Hb$ is the 2D convolution operator, $\Phib$ is a diagonal matrix representing the nonlinear emissivity mapping, and $\rfb$ is the unknown temperature field. We again emphasize that only the operations $\rfb \mapsto \Hb\rfb$ and $\bgb \mapsto \Hb^T\bgb$ are needed in practical implementation of the optimization algorithms. 

\subsection{Infrared image super-resolution}

In classical image super-resolution, the forward model is
\beq
\bgb = \Hb \Db \rfb + \epsilonb,
\label{eq34}
\eeq
where $\rfb$ is the high-resolution (HR) image, $\Db$ is the downsampling operator, $\Hb$ accounts for optical filtering or sensor blur, and $\bgb$ is the observed low-resolution (LR) image.

In the IR setting, we incorporate the same emissivity nonlinearity as in restoration:
\beq
\bgb = \Hb \Db \Phib \rfb + \epsilonb.
\label{eq35}
\eeq
Thus, the super-resolution problem can be regarded as the same forward model as restoration, with an additional downsampling step before final measurement.

\subsection{Implementation details}

Because the forward models of restoration and super-resolution differ only by the presence of $\Db$, their BPINN implementations follow nearly identical steps:

\bit
\item \textbf{Generation of synthetic datasets.}  
We generate a training dataset 
$\{\rfb_{Ti}, \bgb_{Ti}\}$  
using the forward model (\ref{eq33}) or (\ref{eq35}).  
This includes simulating emissivity variations, PSF convolution, downsampling (for SR), and controlled noise.

\item \textbf{Neural network architecture.}  
We first validated all components of the BPINN on a simple fully connected NN with three hidden layers and ReLU activations.  
For realistic imaging, we used a U-Net architecture.  
More advanced architectures (unrolled networks, operator decompositions, plug-and-play variants, etc.) may also be employed depending on the task; see \cite{psf2023009014,proceedings2019033016} for discussion.

\item \textbf{Loss functions and optimization.}  
Training follows the Bayesian framework developed in Section~4, combining  
\[
J(\rwb) = J_{NN}(\rwb) + J_{PI}(\rwb) + J_{PR}(\rwb),
\]  
with the supervised and unsupervised variants implemented exactly as described in Algorithms~\ref{alg:BPINN-IP} and \ref{alg:BPINN_inference}.  
We use gradient-based optimization (Adam) with tuned learning rates.

\item \textbf{Performance monitoring.}  
During training, validation losses, PSNR, and SSIM are tracked across epochs.

\item \textbf{Model saving and reuse.}  
The optimized parameters $\rwbh$ are saved and reloaded for inference and Monte Carlo dropout–based UQ.

\item \textbf{Evaluation on simulated and real IR data.}  
Finally, trained models are tested on synthetic datasets and on real industrial IR images (with controlled emissivity and acquisition conditions).
\eit

A Python Jupyter notebook implementing all steps is available and will be publicly released upon acceptance of the paper.

Figures~\ref{fig02}, \ref{fig03}, and \ref{fig04} show representative examples of the synthetic training, validation, and test datasets.

\vspace{-3mm}
\begin{figure}[h!]
\begin{center}
\includegraphics[scale=.5]{Training_dataset}
\end{center}
\caption{Examples of synthetic IR training data. Top: true temperature images $\rfb_{Ti}$. Bottom: blurred and noisy observations $\bgb_{Ti}$.}
\label{fig02}
\end{figure} 

\newpage
\begin{figure}[h!]
\begin{center}
\includegraphics[scale=.5]{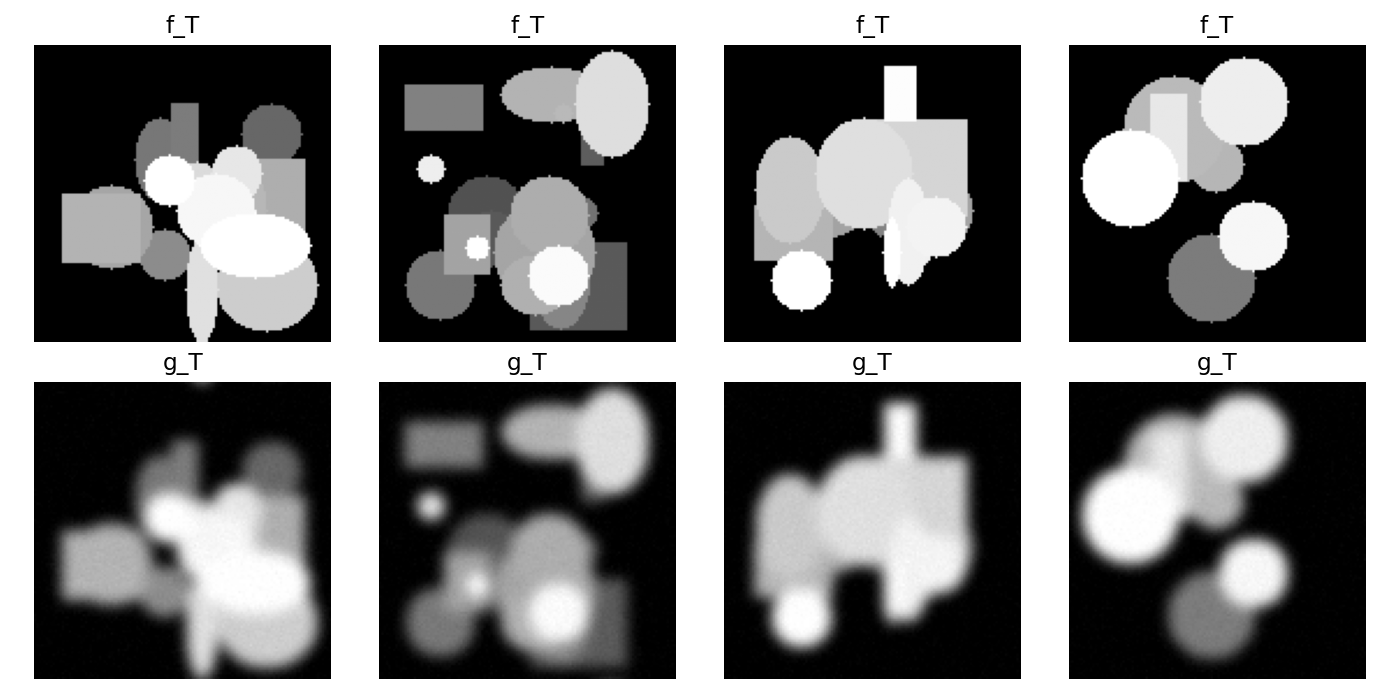}
\end{center}
\caption{Examples of validation images used during training. Top: true images; Bottom: blurred and noisy observations.}
\label{fig03}
\end{figure} 

\begin{figure}[h!]
\begin{center}
\includegraphics[scale=.5]{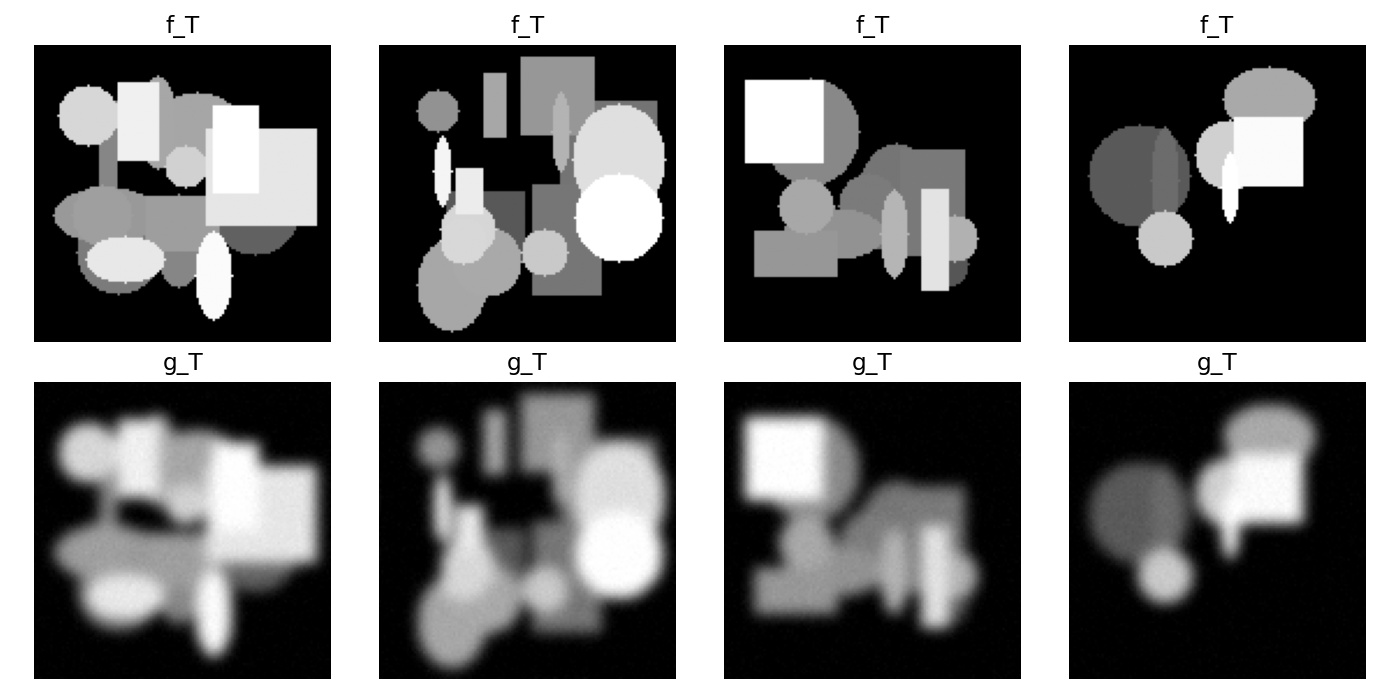}
\end{center}
\caption{Examples of test images. Top: ground truth; Bottom: degraded inputs.}
\label{fig04}
\end{figure} 

\begin{figure}[h!]
\begin{center}
\includegraphics[scale=.5]{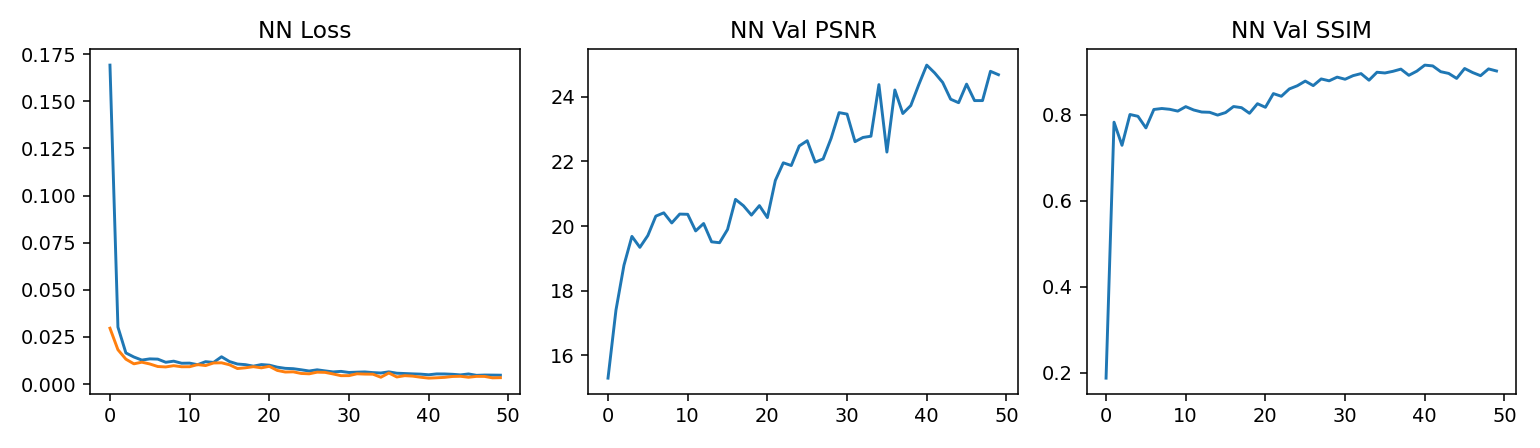}
\end{center}
\caption{Training and validation losses, PSNR, and SSIM for the simple NN model.}
\label{fig05}
\end{figure}

\begin{figure}[h!]
\begin{center}
\includegraphics[scale=.5]{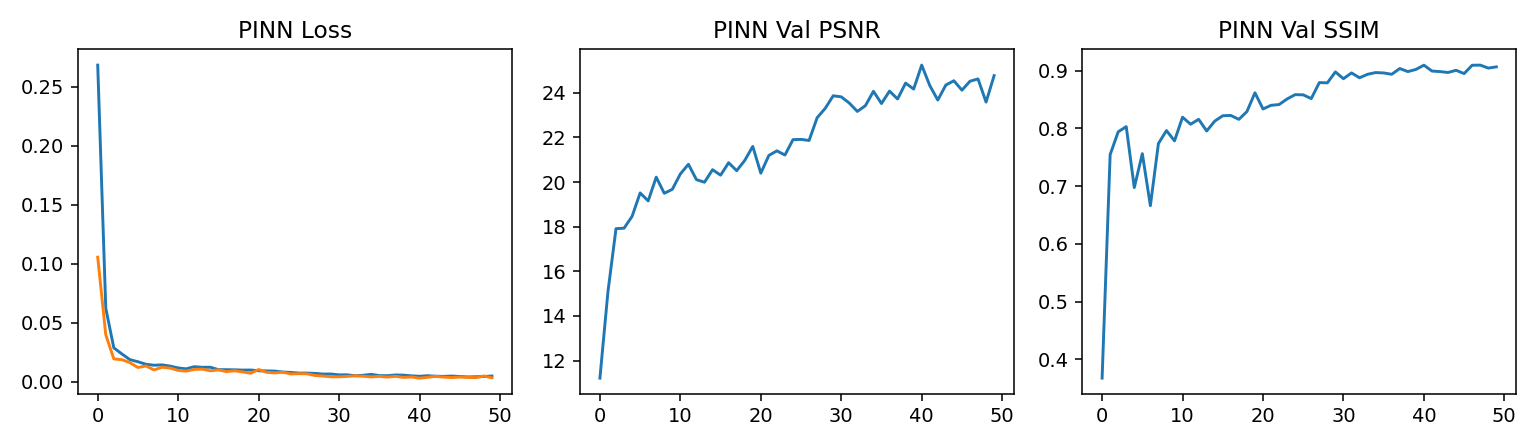}
\end{center}
\caption{Training and validation losses, PSNR, and SSIM for the PINN model.}
\label{fig06}
\end{figure}

\begin{figure}[h!]
\begin{center}
\includegraphics[scale=.5]{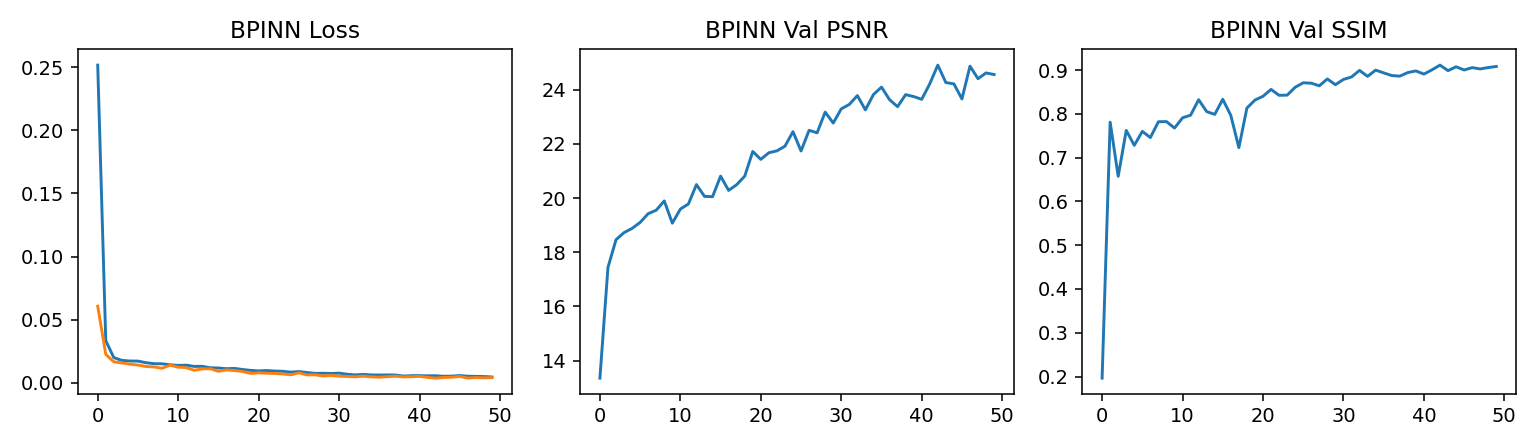}
\end{center}
\caption{Training and validation losses, PSNR, and SSIM for the BPINN-IP model.}
\label{fig07}
\end{figure}

\newpage
Figures~\ref{fig05}, \ref{fig06}, and \ref{fig07} show the evolution of the loss functions, PSNR, and SSIM during training for the NN, PINN, and BPINN, respectively.

\vspace{-2mm}

\vspace{-2mm}
\vspace{-1cm}

\begin{figure}[h!]
\begin{center}
\includegraphics[width=\textwidth,height=16cm]{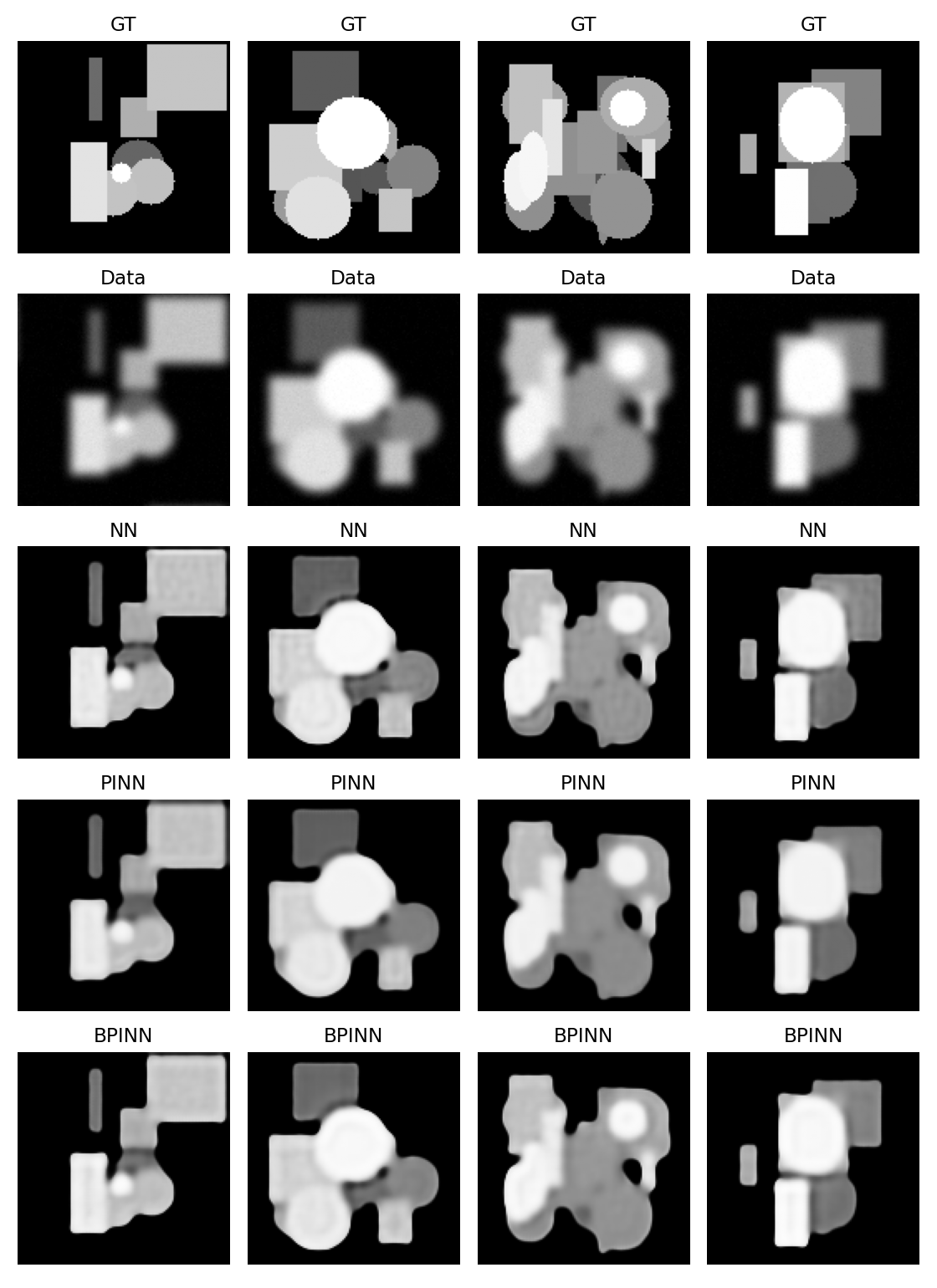}
\end{center}
\caption{Test examples: ground truth, degraded input, and outputs of NN, PINN, and BPINN-IP models.}
\label{fig08}
\end{figure}

\newpage
Figure~\ref{fig08} displays representative test results comparing NN, PINN, and BPINN-IP reconstructions for IR image restoration.

\begin{figure}[h!]
\begin{center}
\includegraphics[scale=.6]{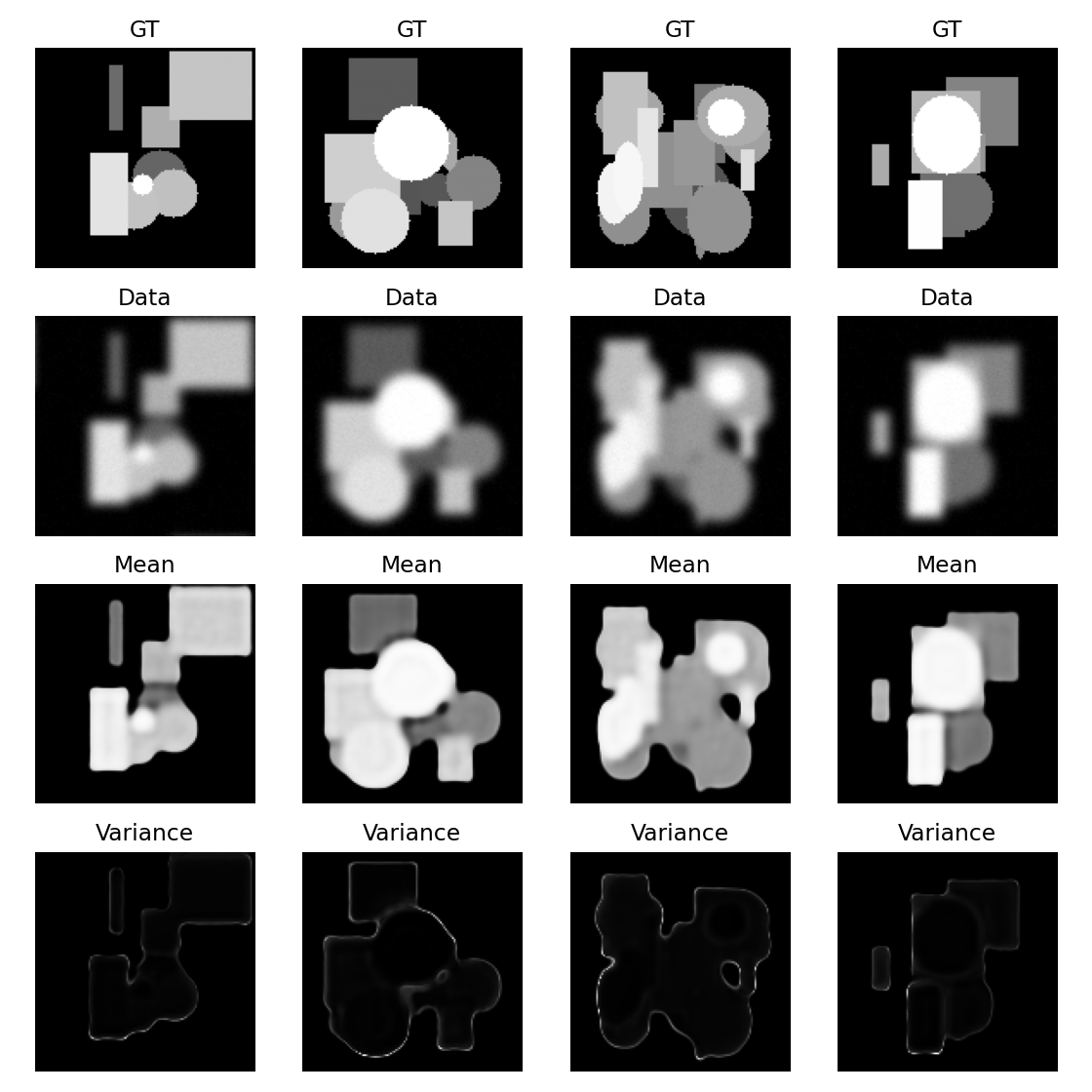}
\end{center}
\caption{BPINN-IP inference results: posterior mean and variance images obtained via MC-dropout (approximate Bayesian Monte Carlo).}
\label{fig09}
\end{figure}

\newpage
Figure~\ref{fig09} shows the posterior mean and variance images produced by the BPINN-IP using Monte Carlo dropout (Algorithm~\ref{alg:BPINN_inference}), illustrating uncertainty quantification.

We used $N_{train}$ synthetic images of size $128 \times 128$ for training, and $N_{val}$ for validation and $N_{test}$ for testing. We simulated two cases, one with great number of images (512, 128, 128) and one (128, 32, 32) respectively for $N_{train}$, $N_{val}$ and $N_{test}$. 

For both restoration and super-resolution, two neural network structures are used: one a CNN with a small number of layers, and the second a more deeper based on UNET. Full implementation details, as well as the ablation experiments are provided in the supplementary material containing also the associated notebooks.

\begin{figure}[h!]
\begin{center}
\includegraphics[scale=.6]{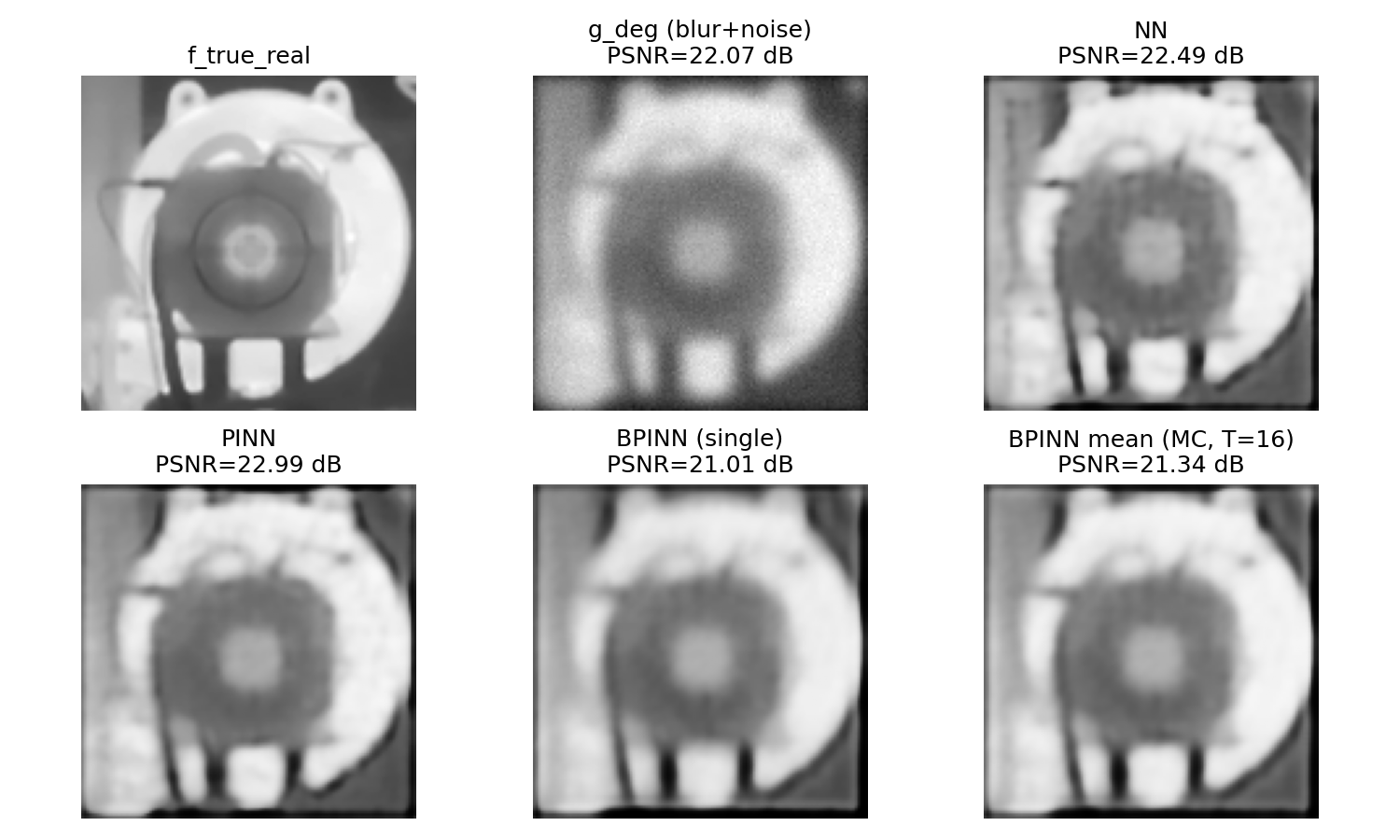}
\end{center}
\caption{A real infrared image (top-left) is degraded (top-midle) and then restored using the three trained models NN, PINN, BPINN-IP single pass and the mean.}
\label{fig10}
\end{figure}

Figure~\ref{fig10} shows an example with a real infrared image which is degraded artificially and then restored using the three trained models NN, PINN, BPINN-IP. 

\begin{figure}[h!]
\begin{center}
\includegraphics[scale=.5]{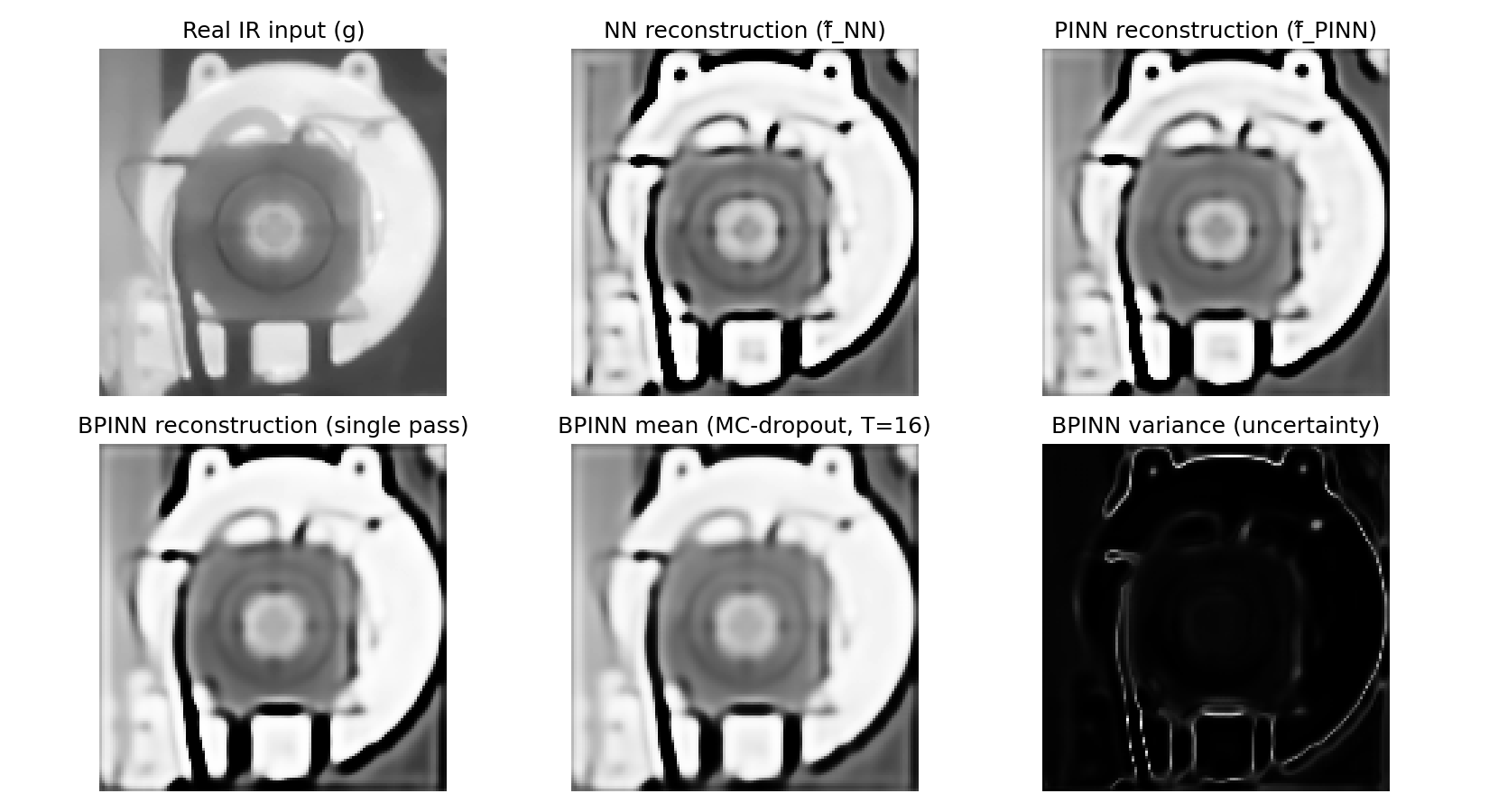}
\end{center}
\caption{Results of the trained super-resolution networks NN, PINN, and BPINN-IP on a real IR imag.}
\label{fig11}
\end{figure}

Figure~\ref{fig11} illustrates the results of the trained super-resolution networks NN, PINN, and BPINN-IP on a real IR image.

Many other examples are given in the supplementary document. 

\newpage
\section{Discussion and Conclusion}

This work introduced a new Bayesian formulation of Physics-Informed Neural Networks (BPINNs) for inverse problems (BPINN-IP) with specific application focus on infrared (IR) image restoration and super-resolution. The framework unifies classical regularization, Bayesian inference, and physics-informed deep learning into a single coherent probabilistic model. 

From a methodological perspective, the key idea is to explicitly model the data–generation process using the forward operator $\Hb$, possibly the nonlinear emissivity mapping $\Phib$, and the associated measurement uncertainties. This generative viewpoint leads systematically to the Bayesian posterior distribution of the unknown image $\rfb$ and, simultaneously, to the posterior distribution of the neural network parameters $\rwb$. Classical PINNs naturally emerge as the Maximum A Posteriori (MAP) solution of the BPINN-IP, while the full Bayesian model provides uncertainty quantification through posterior variances.

Several important observations can be drawn from the results:

\begin{itemize}
\item \textbf{Physical consistency improves reconstruction quality.} \\  
Including the forward operator $\Hb$ in the training loss significantly stabilizes the inverse mapping, especially in the unsupervised setting where ground-truth images are unavailable. 

\item \textbf{The BPINN-IP posterior explains the role of loss terms.}\\ 
The decomposition
\[
J(\rwb) = J_{NN}(\rwb) + J_{PI}(\rwb) + J_{PR}(\rwb)
\]
appears naturally from Bayes’ theorem. The supervised term $J_{NN}$ enforces consistency with labeled data, the physics-informed term $J_{PI}$ enforces fidelity to the forward operator, and the prior term $J_{PR}$ regularizes the NN parameters. This probabilistic interpretation provides a precise justification for the structure of PINN losses used in the literature.

\item \textbf{Bayesian modeling provides principled uncertainty quantification.} \\ 
By interpreting dropout as an approximate Bayesian sampling method, BPINN-IPs provide pixelwise variance maps that identify regions of uncertainty. 

\item \textbf{BPINNs perform well on both synthetic and real IR data.} \\  
The experiments confirm that the method is also robust to nonlinear emissivity effects, spatial blur induced by heat diffusion, and measurement noise. The use of a U-Net backbone further improves performance, showing that the BPINN-IP formulation can be combined with modern architectures.
\end{itemize}

Overall, the proposed BPINN-IP framework provides a general, interpretable, and computationally efficient methodology for solving inverse problems governed by physical models. It bridges the gap between classical Bayesian inference and physics-informed deep learning and offers a practical way to handle large-scale imaging problems where direct Bayesian computation is infeasible.

Future directions include: 
\ben 
\item extending the approach to nonlinear partial differential equations,  
\item exploring more accurate emissivity–radiation models for IR thermography,  
\item incorporating spatially varying PSFs and turbulence effects, and  
\item applying BPINNs to multimodal fusion problems combining IR, visible, and depth measurements.
\een

\newpage
\section{Conclusion}
In this paper, we introduced a new point of view on the Bayesian Physics-Informed Neural Network (BPINN), specifically focusing on inverse problems (BPINN-IP). Even if the generic abbreviation BPINN has been used before for ODEs and PDEs, there was no sign of PINN and BPINN for the explicite inverse problems, where the forward model is a linear one, such as a convolution or super-resolution. Almost all the existing litterature concerning PINN are for ODEs and PDEs. 
In this paper, we focussed on a more general integral linear inverse problems as in equation~[\ref{eq01}] or [\ref{eq32}], and precisely for the discretized vector-matrix presentation as in equation~[\ref{eq02}]  or [\ref{eq33}]. 

By integrating Bayesian inference into the PINN structure, we achieved, not only more robust parameter estimation, but also meaningful uncertainty quantification, which is crucial for some real-world applications with noisy and incomplete data. Of course, the Bayesian approach has been used  in different ways in NN based methods since many years. See for example \cite{Seeger2004,Blundell2015,Kendall2017,Sun2019,Rudner2023,Wenzel2020}. However, the proposed Bayesian approach in PINN is new, as we use a probabilist modeling in the generation or acquisition of the data set used forthe training using the forward model and accountion for the errors. 
In fact the Physics is used both in the training data set generation, and in training of the NN. The obtained posterior probabilities can be used for the training part and for the testing or inference part. 

For the application, our specific focus was on infrared image processing tasks like image deconvolution or super-resolution. Through the realistic simulation studies and a few real industrial examples, we demonstrated that BPINN-IP outperforms traditional methods in robustness and generalisation property of the trained NN, while reducing reliance on large labeled datasets. The ability to encode prior knowledge and physical constraints directly into the learning process significantly enhances generalization capabilities, especially in ill-posed and high-dimensional settings.

Future work will explore extending the BPINN framework to handle non-linear and dynamic inverse problems, optimizing network architectures for faster convergence, and validating the method on larger and more complex real-world data sets. With these improvements, we believe BPINN-IP can become a powerful, practical tool for solving a wide range of inverse problems across science and engineering.

\section*{Acknowledgments}

\bibliographystyle{elsarticle-num}  

\bibliography{references}


\end{document}